\documentclass{clv3}

\usepackage{hyperref}
\usepackage{xcolor}
\definecolor{darkblue}{rgb}{0, 0, 0.5}
\hypersetup{colorlinks=true,citecolor=darkblue, linkcolor=darkblue, urlcolor=darkblue}

\usepackage[export]{adjustbox}

\usepackage{times}
\usepackage{latexsym}

\usepackage{multirow}
\usepackage{url}

\usepackage{wrapfig}
\usepackage{lipsum}
\usepackage{bbm}
\usepackage{array,multirow,graphicx,xcolor}
\usepackage[normalem]{ulem}
\usepackage{amsmath}
\usepackage{amsfonts}
\usepackage{gensymb}
\usepackage{lipsum}
\usepackage{tabularx}
\usepackage{booktabs}
\usepackage{soul}

\usepackage{ragged2e}
\usepackage{float}

\graphicspath{{figures/pdf/},{figures/GMs/}}

\DeclareGraphicsExtensions{.pdf}
\usepackage{tikz}

\usetikzlibrary{bayesnet}

\definecolor{greener}{rgb}{0.0, 0.65, 0.47}

\usepackage[font=small]{caption}
\usepackage{array}
\newcolumntype{L}{>{\centering\arraybackslash}m{3cm}}

\newcolumntype{Y}{>{\RaggedRight\arraybackslash}X} 

\DeclareMathOperator*{\argmax}{arg\,max}
\usepackage{xcolor}

\issue{1}{1}{2016}

\begin{document}

\title{Discourse in Multimedia: A Case Study in Information Extraction}

\author{Mrinmaya Sachan}
\affil{Machine Learning Department\\School of Computer Science\\Carnegie Mellon University}

\author{Kumar Avinava Dubey}
\affil{Machine Learning Department\\School of Computer Science\\Carnegie Mellon University}

\author{Eduard H. Hovy}
\affil{Language Technologies Institute\\School of Computer Science\\Carnegie Mellon University}

\author{Tom M. Mitchell}
\affil{Machine Learning Department\\School of Computer Science\\Carnegie Mellon University}

\author{Dan Roth}
\affil{Department of Computer and Information Science\\University of Pennsylvania}

\author{Eric P. Xing}
\affil{Machine Learning Department\\School of Computer Science\\Carnegie Mellon University}

\maketitle

\begin{abstract}
To ensure readability, text is often written and presented with due formatting. These text formatting devices help the writer to effectively convey the narrative. At the same time, these help the readers pick up the structure of the discourse and comprehend the conveyed information. There have been a number of linguistic theories on discourse structure of text. However, these theories only consider unformatted text. Multimedia text contains rich formatting features which can be leveraged for various NLP tasks. In this paper, we study some of these discourse features in multimedia text and what communicative function they fulfil in the context. We examine how these multimedia discourse features can be used to improve an information extraction system. We show that the discourse and text layout features provide information that is complementary to lexical semantic information commonly used for information extraction. As a case study, we use these features to harvest structured subject knowledge of geometry from textbooks. We show that the harvested structured knowledge can be used to improve an existing solver for geometry problems, making it more accurate as well as more explainable.
\end{abstract}

\section{Introduction}
The study of discourse focuses on the properties of text as a whole and how meaning is conveyed by making connections between component sentences. Writers often use certain linguistic devices to make a discourse structure which enables them to effectively communicate their narrative. The readers too comprehend text by picking up these linguistic devices and recognizing  the discourse structure. There are a number of linguistic theories on discourse relations \cite{van:1972,Longacre:1983,grosz1986attention,cohen1987analyzing,mann:1988,polanyi1988formal,moser1996toward} which specify relations between discourse units and how to represent the discourse structure of a piece of text i.e. discourse  parsing \cite{duverle2009novel,subba2009effective,feng2012text,ghosh2012global,feng-hirst:2014,ji2014representation,li2014text,lin2014pdtb,Wang2015ARE}. These discourse features have been shown to be useful in a number of NLP applications such as summarization \cite{van1979recalling,marcu:2000,boguraev2000discourse,louis2010discourse,gerani2014abstractive}, information retrieval \cite{wang2006information,lioma2012rhetorical}, information extraction \cite{kitani1994pattern,conrath2014unsupervised} and question answering \cite{chai2004discourse,sun2007discourse,Narasimhan:15,sachan:2015,Sachan:2016b}.

Most linguistic theories of discourse consider written text without much formatting. However, in this multimedia age, text is often richly formatted. Be it newsprint, textbooks, brochures, or even scientific articles, text is usually appropriately formatted and stylized. For example, the text may have a heading. It may be divided into a number of sections with section subtitles. Parts of the text may be italicized or boldfaced to place appropriate emphasis wherever required. The text may contain itemized lists, footnotes, indentations or quotations. It may refer to associated tables and figures. The tables and figures too usually have associated captions. All these text layout features ensure that the text is easy to read and understand. Even articles accepted at the \textit{Computational Linguistics} journal follow a due formatting scheme.

These text layout features are in addition to other linguistic devices such as syntactic arrangement or rhetorical forms. Relations between textual units that are not necessarily contiguous can thus be expressed thanks to typographical or dispositional markers. Such relations, which are out of reach of standard NLP tools, have only been studied within some specific layout contexts \cite[inter alia]{hovy:1998,pascual:1996,Bateman:2001}\footnote{Please see related work (section \ref{rw}) for a complete list of references.} and there are no comprehensive studies on the various kinds of discourse features and how they can be leveraged to improve NLP tasks. Moreover, there are no large scale corpus studies of multimedia text.

\textit{In this paper, we study some of these discourse features in multimedia text and what communicative function they fulfil in the context. We perform the first large scale corpus study of multimedia text and shows that the formatting devices can indeed be used to improve a strong information extraction system. We conclude that the discourse and text layout features provide information that is complementary to lexical semantic information commonly used for information extraction.}

\begin{figure}
	\center
	\includegraphics[scale=0.5]{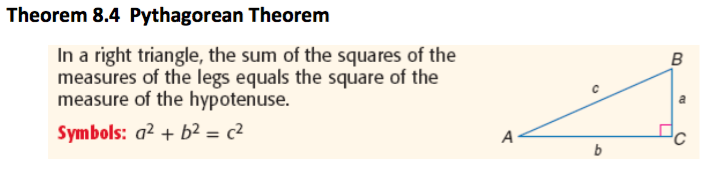}
	\caption{An excerpt of a textbook from our dataset that introduces the Pythagoras theorem. The textbook has a lot of typographical features that can be used to harvest this theorem: The textbook explicitly labels it as a ``theorem''; there is a colored bounding box around it; an equation writes down the rule and there is a supporting figure. Our models leverages such rich contextual and typographical information (when available) to accurately harvest axioms and then parses them to horn-clause rules.}\label{fig:pythagorous}
\end{figure}
As a case study, we study the problem of harvesting structured subject knowledge of geometry from textbooks.
With the intent of making the subject material easy to grasp and remember for students, textbooks often contain rich discourse and formatting features. Crucial material such as axioms or theorems are presented with stylistic highlighting or bounding boxes. Often, mathematical information such as equations are presented in a separate color and font size. Often, theorems are numbered or named (e.g. Theorem 8.4). For example, Figure \ref{fig:pythagorous} shows a snapshot of a math textbook which describes the Pythagoras  theorem. 
The textbook explicitly labels it as a ``theorem''; there is a colored bounding box around it; an equation writes down the rule and there is a supporting figure.

In this paper, we will try to answer the question: \textit{Can this rich contextual and typographical information (whenever available) be used to harvest such axioms in the form of structured rules?} For example, we wish to not only extract the axiom mention in Figure \ref{fig:pythagorous} but also map it to a rule corresponding to the Pythagoras theorem:
\begin{eqnarray*}
	isTriangle(ABC) \land perpendicular(AC, BC) \implies BC^2 + AC^2 = AB^2
\end{eqnarray*}

We propose novel models which perform sequence labeling and alignment to extract redundant axiom mentions across various textbooks, and then parse the redundant axioms to structured rules. These redundant structured rules are then resolved to achieved the best correct structured rule for each axiom. We conduct a comprehensive feature analysis of the usefulness of various discourse features: shallow discourse features based on discourse markers, a deep one based on \textit{Rhetorical Structure Theory} \cite{mann:1988}, and various text layout features in a multimedia document \cite{hovy:1998} for the various stages of information extraction. Our experiments show the usefulness of all the various typographical features over and above the various lexical semantic and discourse level features considered for the task. A shorter version of this paper appeared as \citet{sachan2017textbooks}.

\section{Background and Related Work}\label{rw}
\noindent{\bf Discourse Analysis:} Discourse analysis is the analysis of semantics conveyed by a coherent sequence of sentences, propositions or speech. Discourse analysis is taken up in a variety of disciplines in the humanities and social sciences and a number of discourse theories have been proposed \cite[inter alia]{kamp1993discourse,mann:1988,lascarides2008segmented}. Their starting point lies in the idea that text is not just a collection of sentences, but it also includes relations between all these sentences that ensure its coherence. It is often assumed that discourse analysis is a 3-step process:
\vspace{-0.5cm}\begin{enumerate}
	\item splitting the text into Discourse Units (DU)
	\item ensuring the attachment between DUs, and then
	\item labeling links between DUs with discourse relations.
\end{enumerate}\vspace{-0.5cm}
Discourse relations may be divided into two categories: nucleus-satellite (or subordinate) relations which link an important argument to an argument supporting background information, and multi-nuclear (or coordinate) relations which link arguments of equal importance. Most of discourse theories (DRT, RST, SDRT, etc.) acknowledge that a discourse is hierarchically structured thanks to discourse relations. A number of discourse relations have been proposed under various theories for discourse analysis. 

Discourse analysis has been shown to be useful for many NLP tasks such as question answering \cite{chai:2004,lioma:2012,jansen-surdeanu-clark:2014}, summarization \cite{louis2010} and information extraction \cite{kitani1994pattern}. However, to the best of our knowledge, we do not have a theory or a working model of discourse in a multi-media setting.

{\bf Formatting in Discourse:} 
Psychologists and educationists have frequently studied multimedia issues such as the impact of illustrations (pictures, tables, etc.) in text, design principles of multimedia presentations, etc. \cite{Dwyer:78,fleming1978instructional,hartley1985designing,twyman1985using}. However, these discussions are usually too general and hard to build on from a computational perspective. 
Thus, most studies of multimedia text have only been theoretical in nature. \citet{larkin1987diagram,mayer1989systematic,petre:90} attempt to answer questions: whether a graphical notation is superior to text notation, what makes a diagram (sometimes) worth ten thousand words, how illustration effects thinking. 
\citet{hovy:1998,arens1990describe,Arens:92,arens1993structure} provide a theory of the communicative function fulfilled by various formatting devices and use it in text planning. In a similar vein, \citet{dale1991role,dale1991exploring,White:95,pascual1996semantic,reed1997generating,bateman2001towards} discuss the textual function of punctuation marks and use it in the text generation process. \citet{andre1991wip,andre2000generation} build a system \textit{WIP} that generates multimedia presentations via layered architecture (comprising of the control layer, content layer, design layer, realization layer and the presentation layer) and with the help of various content, design, user and application experts.
 \citet{mackinlay1986automating} discuss the automatic generation of tables and charts. \citet{luc1999linguistic} study enumerations.  \citet{feiner1988architecture,arens1988automatic,neal1990intelligent,feiner1991automating,wahlster1992wip,arens1992knowledge,maybury1998planning} discuss various aspects of processing and knowledge required for automatically generating multimedia. Finally, \citet{stock1993alfresco} discuss using hypermedia features for the task of information exploration.

However, all the aforementioned studies were merely theoretical. All the models were hand-coded and not trained from multimedia corpora. In this paper, we provide the first corpus analysis of multimedia text and use it to show that the formatting devices can indeed be used to improve a strong information extraction system.

{\bf Solving Geometry Problems:}
% Standardized tests have been recently proposed as `drivers for progress in AI' \cite{Clark:2016mag}. These tests are easily accessible, and measurable, and hence have attracted several NLP researchers. There is a growing body of work on solving standardized tests such as reading comprehensions \cite[inter alia]{richardson:2013}, science question answering \cite[inter alia]{Schoenick:2016}, algebra word problems \cite[inter alia]{kushman:2014}, geometry problems \cite{seo:2015} and pre-university entrance exams \cite{fujita:2014}.
While the problem of using computers to solve geometry questions is old \cite{feigenbaum:1963,Schattschneider:97,davis:2006}, NLP and computer vision techniques were first used to solve geometry problems in \citet{seo:2015}. While \citet{seo:2014} only aligned geometric shapes with their textual mentions, \citet{seo:2015} also extracted geometric relations and built \textit{GEOS}, the first automated system to solve SAT style geometry questions. \textit{GEOS} used a coordinate geometry based solution by translating each predicate into a set of manually written constraints. A Boolean satisfiability problem posed with these constraints was used to solve the multiple-choice question. \textit{GEOS} had two key issues: (a) it needed access to answer choices which may not always be available for such problems, and (b) it lacked the deductive geometric reasoning used by students to solve these problems.
%, rendering them less useful in educational applications such as MOOCs. 
In this paper, we build an axiomatic solver which mitigates these issues by performing deductive reasoning using axiomatic knowledge extracted from textbooks. Furthermore, we use ideas from discourse to automatically extract these axiom rules from textbooks.
%The close relationship between the fields of geometry and logic dates back to \textit{Euclid} who laid down the postulates of axiomatic geometry in his classic treatise \textit{Elements}. Since then, many other famous mathematicians have significantly contributed to the axiomatic formulation of geometry that we study today. 

Automatic approaches that use logical inference for geometry theorem proving such as the Wus method \cite{wen:1986}, Grobner basis method \cite{kapur:1986}, and angle method \cite{chou:1994} have been used in tutoring systems such as \textit{Geometry Expert} \cite{gao:2002} and \textit{Geometry Explorer} \cite{wilson:2005}. There has also been research in synthesizing geometry constructions given logical constraints \cite{gulwani:2011,itzhaky:2013} or generating geometric proof problems \cite{Alvin:2014} for applications in tutoring systems. Our approach can be used to provide the axiomatic information necessary for these works. In addition, there has been some recent work on building a solver for Newtonian physics problems in the AP physics exam \cite{Sachan:2018c,Sachan:2018a}. We can use the technique presented in this paper to obtain structured knowledge for these solvers.

{\bf Information Extraction from Textbooks:} Our model for extracting structured rules of geometry from textbooks builds upon ideas from Information extraction (IE), which is the task of automatically extracting structured information from unstructured and/or semi-structured documents. While there has been a lot of work in IE on domains such as web documents \cite{chang:2003,Etzioni:2004,Cafarella:2005,chang:2006,Banko:2007,etzioni:2008,Mitchell:2015} and scientific publication data \cite{shah:2003,peng:2006,saleem:2012}, work on IE from educational material is much more sparse.
Most of the research in IE from educational material deals with extracting simple educational concepts \cite{shah:2003,canisius:2007,yang:2015,wang:2015,liang:2015,Wu:2015,liu:2016,wang:2016} or binary relational tuples \cite{balasubramanian:2002,clark:2012,Dalvi:2016} using existing IE techniques. On the other hand, our approach extracts axioms and parses them to horn clause rules. This is much more challenging. Raw application of rule mining or sequence labeling techniques used to extract information from web documents and scientific publications to educational material usually leads to poor results as the amount of redundancy in educational material is lower and the amount of labeled data is sparse. Our approach tackles these issues by making judicious use of typographical information, the redundancy of information and ordering constraints to improve the harvesting and parsing of axioms. This has not been attempted in previous work.

{\bf Language to Programs:} After harvesting axioms from textbooks, we also parse the axiom mentions to horn clause rules.
This work is related to a large body of work on semantic parsing \cite[inter alia]{zelle:1993,zelle:1996,kate:2005,zettlemoyer:2012}. Semantic parsers typically map natural language to formal programs such as database queries \cite[inter alia]{liang:2011,Berant:2013,Yaghmazadeh:17}, commands to robots \cite[inter alia]{Shimizu:2009,matuszek:2010,chen:2011}, or even general purpose programs \cite{lei:2013,ling:2016,yin:17,ling:2017}. More specifically, \citet{Chang:2016} and \citet{Quirk:15} learn ``If-Then'' and ``If-This-Then-That'' rules, respectively. In theory, these works can be adapted to parse axiom mentions to horn-clause rules. However, this would require a large amount of supervision which would be expensive to obtain. We mitigated this issue by using redundant axiom mention extractions from multiple textbooks and then combining the parses obtained from various textbooks to achieve a better final parse for each axiom.

\section{Data Format}
% For example, Table \ref{fig:examples} shows some more excerpts from high-school math textbooks. Each excerpt uses rich typography and stylistics in order to effectively communicate content to students and help them grasp the material better. These include highlighting (bold face, italicization, etc.) of the axiom names and the corresponding section names, using accompanying figures, writing down equations stating the axiom rules, etc. 
\begin{table}\center
	\begin{tabular}{|c|}\hline
		\includegraphics[scale=0.41]{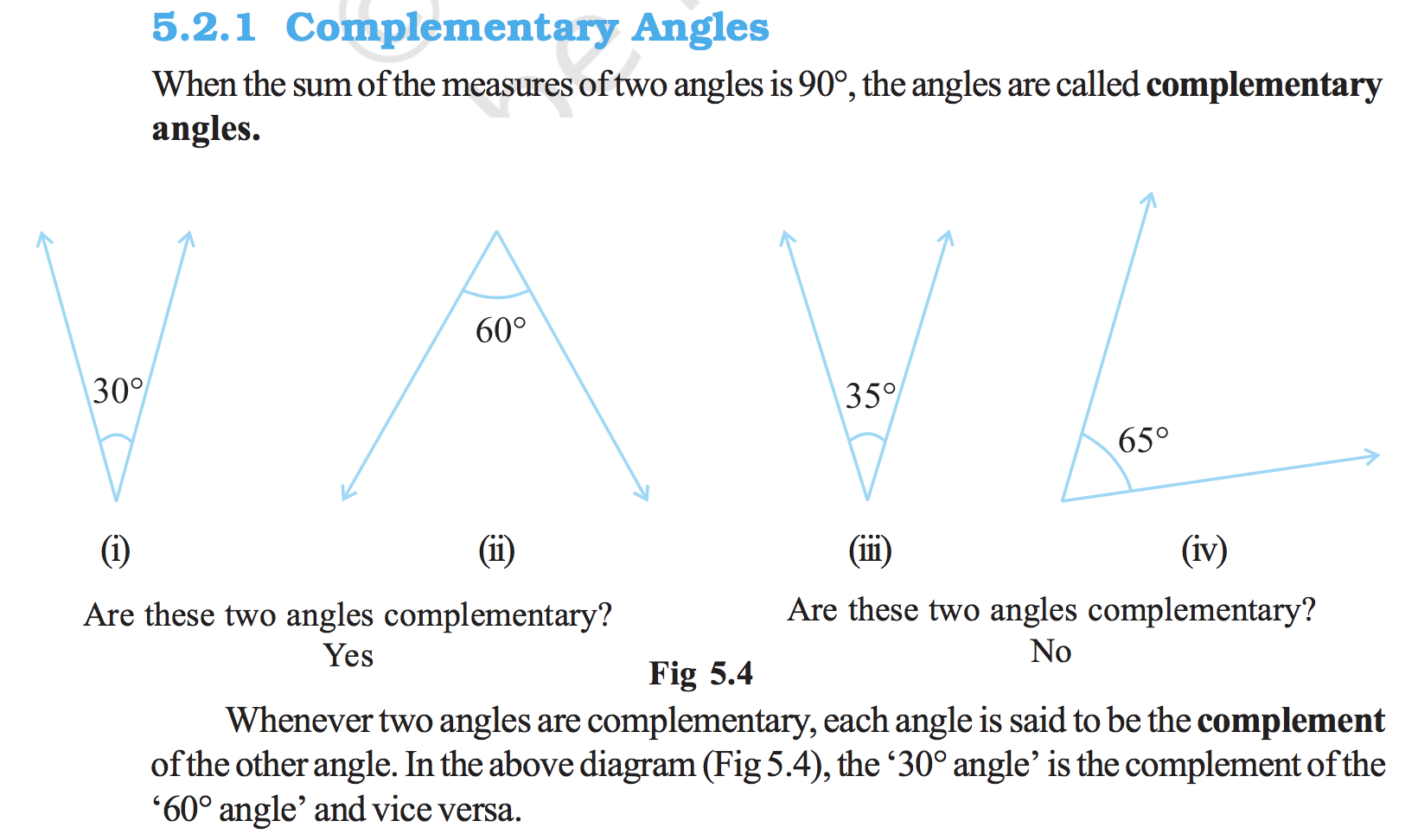}\\\hline
		\includegraphics[scale=0.4]{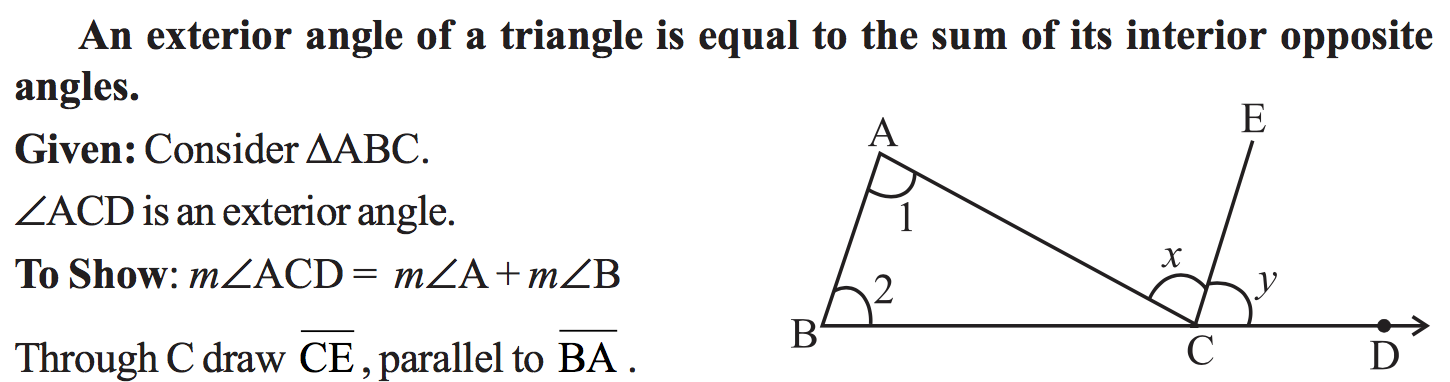}\\\hline
		\includegraphics[scale=0.41]{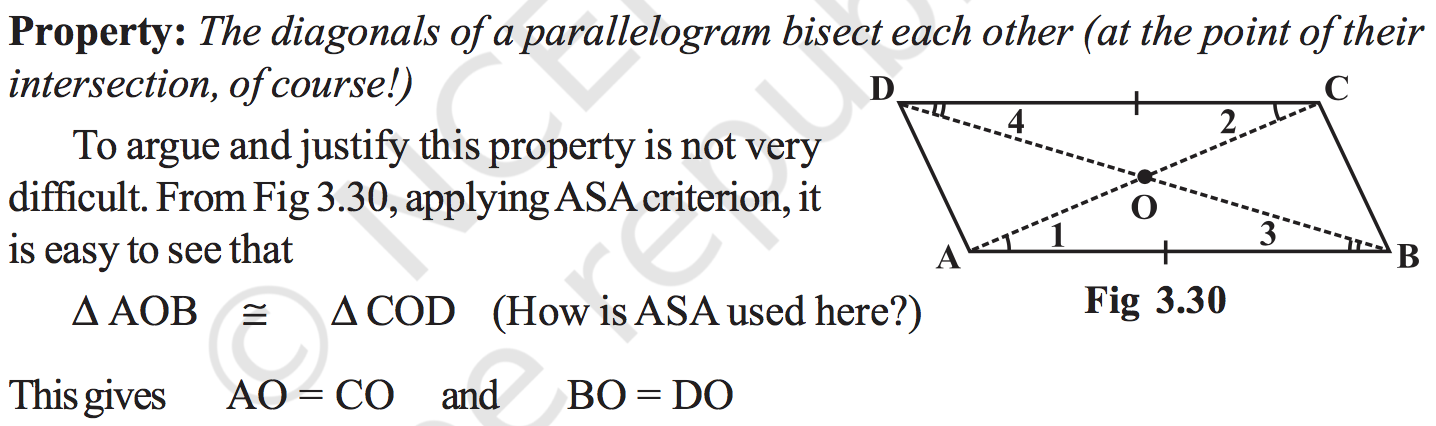}\\\hline
	\end{tabular}\label{book-snippets}
	\caption{Some more excerpts of textbooks from our dataset that describe (a) complementary angles, (b) exterior angles, and (c) parallelogram diagonal bisection axioms. Each excerpt contains rich typographical features that can be used to harvest the axioms. (a) For the complementary angles mention, the textbook explicitly labels the section name ``5.2.1 Complementary Angles'' with bold face and color; the axiom name ``complementary angles'' is in bold font, and there is a supporting figure. (b) For the  exterior angles mention, the axiom statement is bold faced, the axiom rule is mentioned via an equation (which is emphasized with the bold faced string ``To show''), and there is a supporting figure. (c) For the parallelogram diagonal bisection mention, the axiom statement is emphasized with the bold faced string ``Property'', the axiom statement itself is italicized, there is a supporting figure, and the axiom rule is written as an equation.
		Our model will leverage such rich contextual and typographical information (when available) to accurately harvest axioms and then parses them to horn-clause rules.}\label{fig:examples}
\end{table}
\begin{table}\center
	\begin{tabular}{|c|c|}\hline
		\includegraphics[scale=0.37]{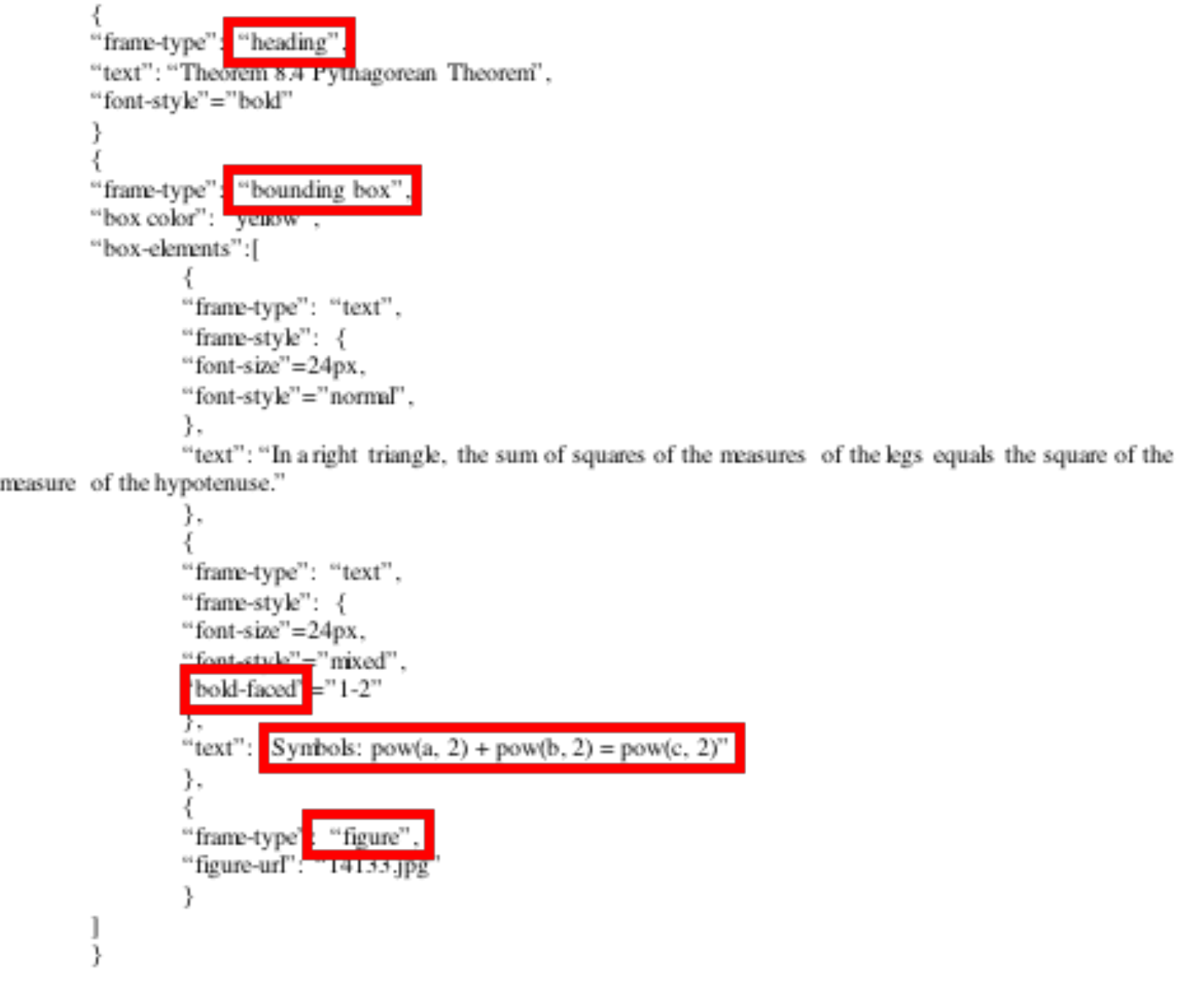}& \includegraphics[scale=0.37]{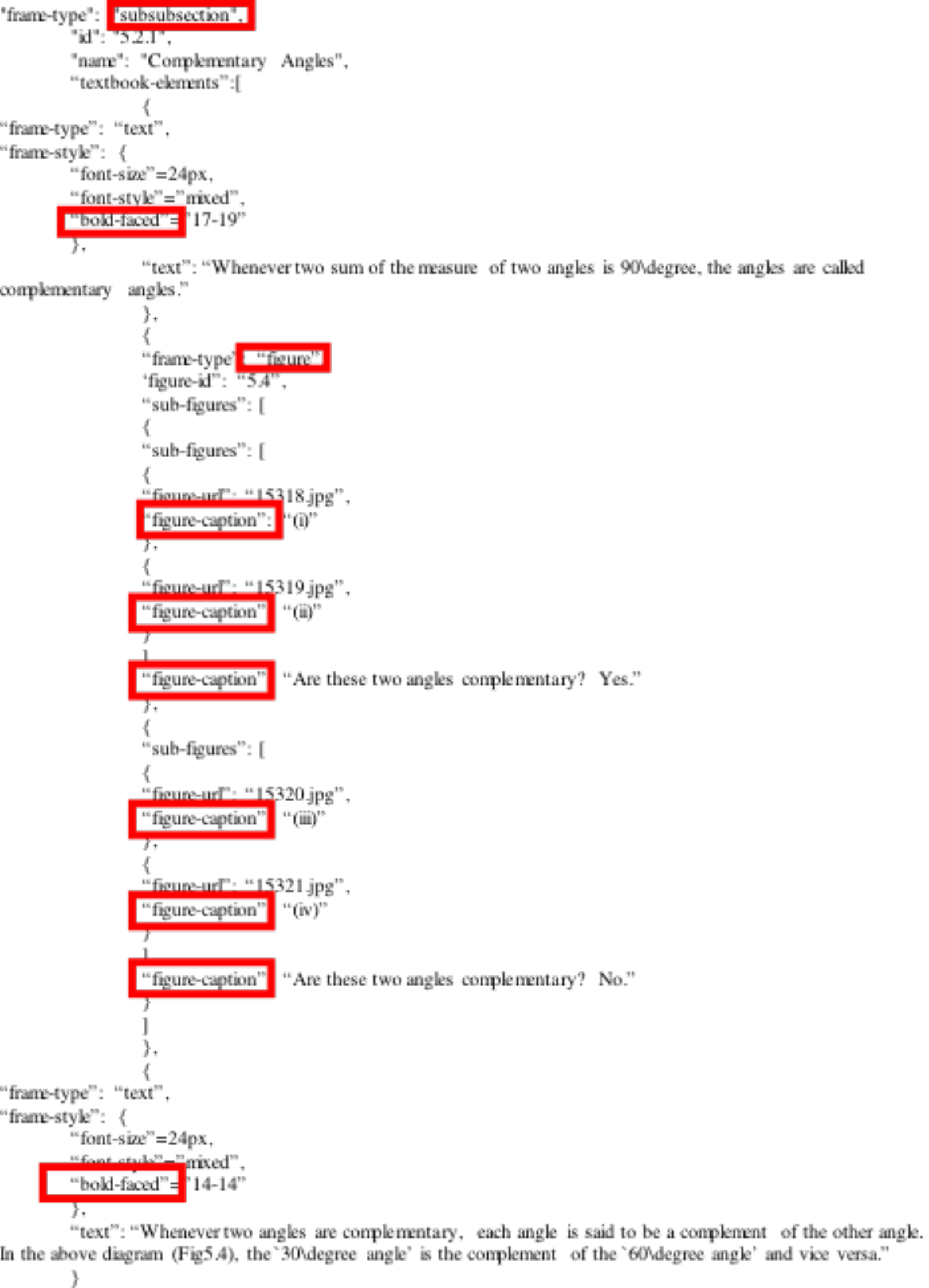}\\\hline
		\includegraphics[scale=0.37]{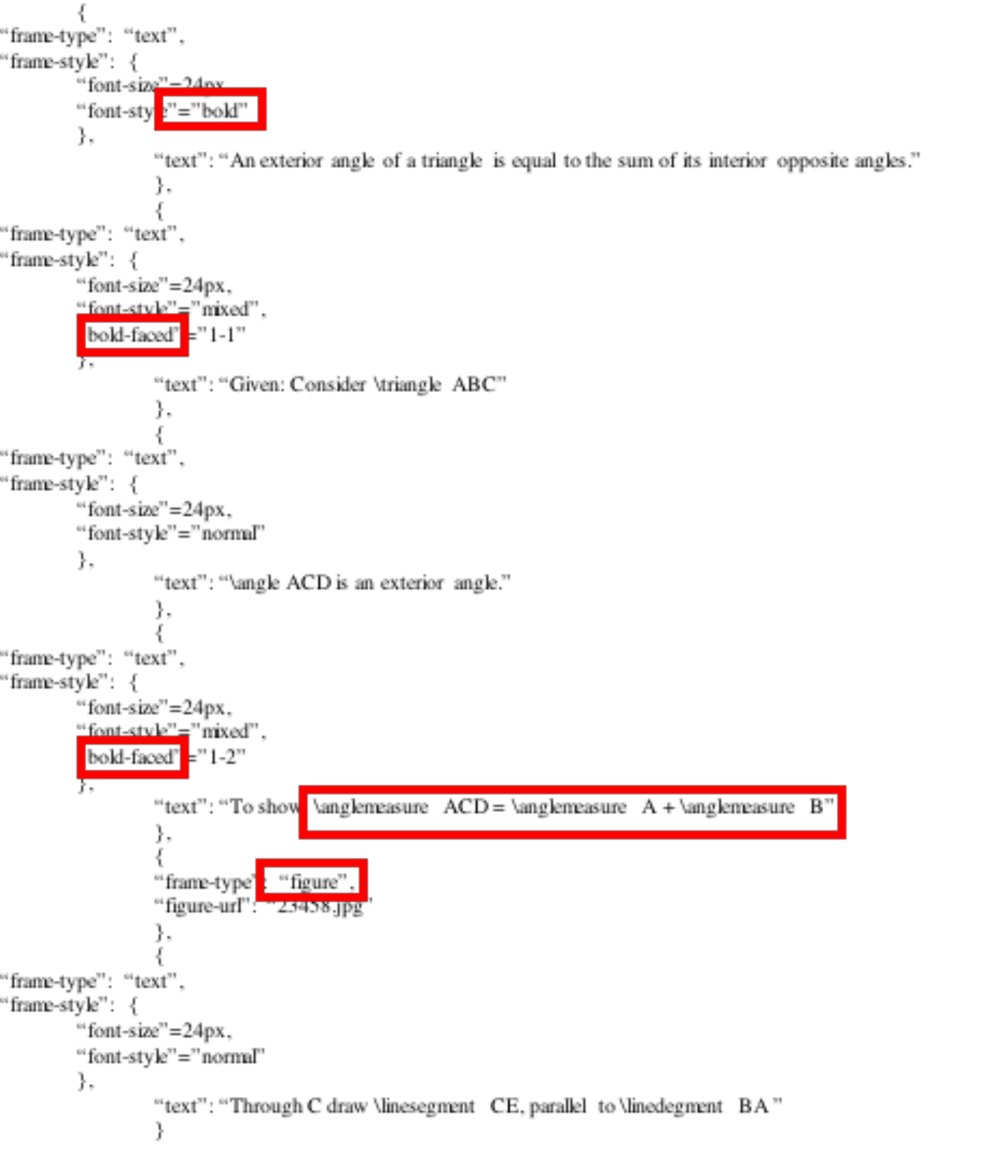}&\includegraphics[scale=0.37]{fig2_json.pdf} 	~\\\hline
	\end{tabular}\label{book-jsons}
	\caption{Corresponding json files for the example textbook excerpt shown in Figure \ref{fig:pythagorous} and the three example textbook excerpts shown in Table \ref{fig:examples}. We mark the various typographical features that can be used to harvest the axioms in red: (a) For the Pythagoras theorem mention, we have features such as the heading, the bounding box, a supporting figure and the equation, (b) For the  complementary angles mention, we have features such as the subsubsection ``5.2.1 Complementary Angles'' with bold face and color; the axiom name ``complementary angles'' is in bold font, and there is a supporting figure. (c) For the  exterior angles mention, the axiom statement is bold faced, the axiom rule is mentioned via an equation (which is emphasized with the bold faced string ``To show''), and there is a supporting figure. (d) For the parallelogram diagonal bisection mention, the axiom statement is emphasized with the bold faced string ``Property'', the axiom statement itself is italicized, there is a supporting figure, and the axiom rule is written as an equation. Our model will leverage these rich contextual and typographical information to harvest axioms and then parses them to horn-clause rules.}\label{fig:examples_json}
\end{table}
Large scale corpus studies of multimedia text have been rare because of the difficulty in obtaining rich multimedia documents in analyzable data structures.
A large proportion of text today is typeset using some typesetting software such as Latex, Word, HTML, etc. These features can also serve as useful cues in downstream applications and a model for text formatting is required.

Table \ref{fig:examples} shows some excerpts of textbooks from our dataset that describe complementary angles, exterior angles, and parallelogram diagonal bisection axioms. As described, each excerpt contains rich typographical features such as the section headings, italicization, bold face, coloring, explicit axiom name, supporting figures, and equations that can be used to harvest the axioms. We wish to leverage such rich contextual and typographical information to accurately harvest axioms and then parses them to horn-clause rules. The textbooks are provided to us in rich json format which retains the rich typesetting of these textbooks as shown in Table \ref{fig:examples_json}. For demonstration, we have manually marked the various typographical features that can be used to harvest the axioms. We will show how we can use these features to harvest axioms of geometry from textbooks and then parse them to structured rules.

\section{Text Formatting Elements in Discourse}
In this section, we review various text formatting devices used in a typical multimedia system and what communicative function do they serve. This will help us come up with a theory for text formatting in discourse and also motivate how these features can be used in a typical NLP application like information extraction. This theory is inspired from various style suggestions for English writing \cite{strunk2007elements}.
The goal of a text formatting device in a multimedia text is to delimit the portion of text for which certain exceptional conditions of interpretation hold. We categorize text formatting devices into four broad categories: \textit{depiction}, \textit{position}, \textit{composition} and \textit{substantiation}, and describe the various text formatting devices below:

\vspace{-0.5cm}
\begin{itemize}
	\item {\bf Depiction:} Depiction features concern with how a string of text is presented in the multimedia. These include features such as capitalization, font size/color, boldface, italicization, underline, strikethrough, parenthesis, quotation marks, use of bounding boxes, etc.
	\item {\bf Position:} Position features concern with the positioning of a piece of text relative to the remaining material in the document. These features include in lining, text offset, footnotes, headers and footers, text separation or isolation (a block of text separated from the rest to create a special effect), 
	\item {\bf Composition:} Composition features concern with the internal structuring of a piece of text. Examples include  graphical markers such as paragraph	breaks, sections (having sections, chapters, etc. in the document), lists (itemization, enumeration), concept definition using a parenthesis or colon, etc. 
	\item {\bf Substantiation:} Substantiation features are used to further substantiate the discourse argument. Examples include associated figures or tables, references to tables, figures (for example, Figure 1.2) or external links which are very important in understanding a complex multimedia document.
\end{itemize}
\vspace{-0.5cm}

\section{The Question}
A key question for research is: ``Are these text formatting features useful for NLP tasks?'' In particular, in this paper, we will try to answer if these text formatting features are useful for information extraction. In a typical multimedia document, authors use various text formatting devices to better communicate the content to their readers. This helps the readers digest the material quickly and much more easily. Thus, can these text formatting features be useful in an information extraction system too? We experimentally validate our hypothesis in the application of harvesting axioms of geometry from richly formatted textbooks.

\begin{figure}
	\center
	\includegraphics[scale=0.35]{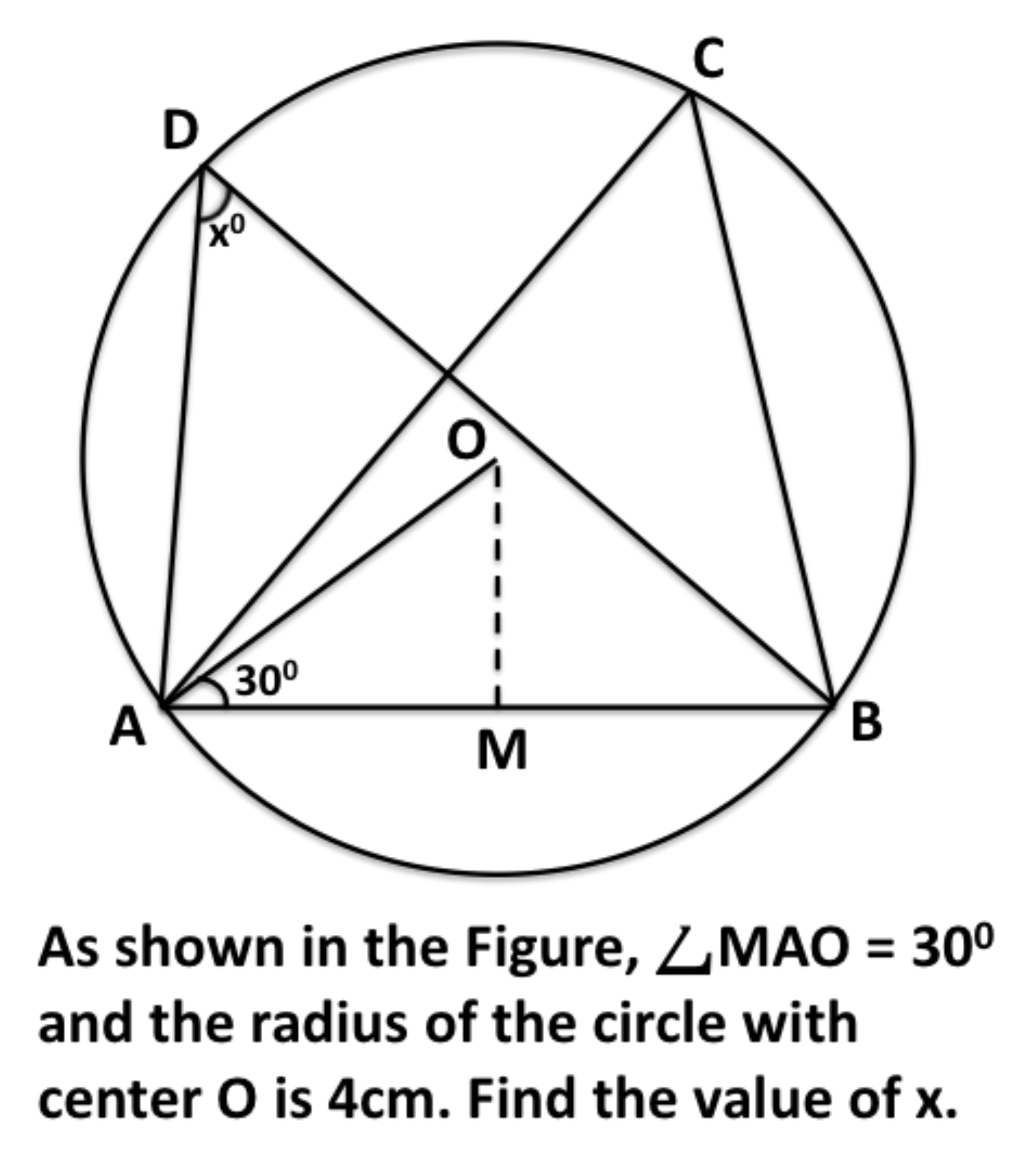}
	\caption{An example SAT style geometry problem. The problem consists of a diagram as well as the question text. In order to solve such a question, the system is required to understand both the diagram as well as the question text, and also reason about geometrical concepts using well-known axioms of Euclidean geometry.}\label{fig:geo-que}
\end{figure}
Then, we show that these harvested axioms can improve an existing solver for answering SAT style geometry problems. SAT geometry tests the student's knowledge of Euclidean geometry in its classical sense, including the study of points, lines, planes, angles, triangles, congruence, similarity, solid figures, circles, and analytical geometry. A typical geometry problem is provided in Figure \ref{fig:geo-que}. Geometry questions includes a textual description accompanied by a diagram. Various levels of understanding are required to solve geometry problems. An important challenge is understanding both the diagram (which consists of identifying visual elements in the diagram, their locations, their geometric properties, etc.) and the text simultaneously, and then reasoning about the geometrical concepts using well-known axioms of Euclidean geometry.

%\section{Application: Harvesting Geometry Axioms from Math Textbooks}

%\textit{GEOS} parses the question text and the diagram to a formal problem description. \textit{GEOS} uses a logical formula, a first-order logic expression (see Figure \ref{fig:example}) that includes known numbers or geometrical entities (e.g. 4 cm) as constants, unknown numbers or geometrical entities (e.g. O) as variables, geometric or arithmetic relations (e.g. \textit{isLine}, \textit{isTriangle}) as predicates and properties of geometrical entities (e.g. \textit{measure}, \textit{liesOn}) as functions. This is done by learning a set of relations that potentially correspond to the question text (or the diagram) along with a confidence score.

We first recap \textit{GEOS}, a completely automatic solver for geometry problems. We will then use the rich contextual and typographical information in textbooks to extract structured knowledge of geometry. This structured knowledge of geometry will then be used to improve \textit{GEOS}.

\section{Background: GEOS}\label{sec:background}

\begin{figure}
	\includegraphics[scale=0.25]{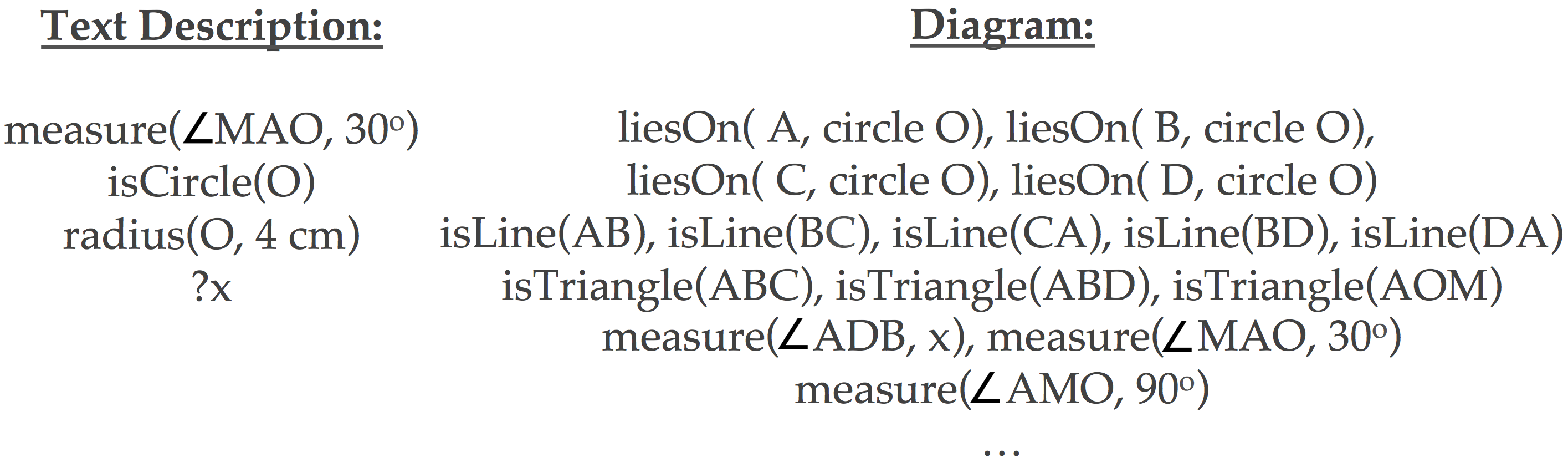}
	\caption{A logical expression that represents the meaning of the text description and the diagram in the geometry problem in Figure \ref{fig:geo-que}. \textit{GEOS} derives a weighted logical expression where each predicates also carries a weighted score but we do not show them here for clarity.}\label{fig:example}
\end{figure}

Our work reuses \textit{GEOS} \cite{seo:2015} to parse the question text and diagram into its formal problem description as shown in Figure \ref{fig:example}. %We first recap \textit{GEOS}. 
\textit{GEOS} uses a logical formula, a first-order logic expression that includes known numbers or geometrical entities (e.g. 4 cm) as constants, unknown numbers or geometrical entities (e.g. O) as variables, geometric or arithmetic relations (e.g. \textit{isLine}, \textit{isTriangle}) as predicates and properties of geometrical entities (e.g. \textit{measure}, \textit{liesOn}) as functions.

This is done by learning a set of relations that potentially correspond to the question text (or the diagram) along with a confidence score. %Then, a subset of the relations that maximize the joint text and diagram scores are picked as the formal problem description.
For diagram parsing, \textit{GEOS} uses a publicly available diagram parser for geometry problems \cite{seo:2014} to obtain the set of all visual elements, their coordinates, their relationships in the diagram, and their alignment with entity references in the question text. The diagram parser also provides confidence scores for each literal to be true in the diagram.
For text parsing, \textit{GEOS} takes a multi-stage approach, which maps words or phrases in the text to their corresponding concepts, and then identifies relations between identified concepts.
% Finally, it performs \textit{relation completion} which handles implications and coordinating conjunctions.

Given this formal problem description,  \textit{GEOS} use a numerical method to check the satisfiablity of literals by defining a relaxed indicator function for each literal. These indicator functions are manually engineered for every predicate. 
Each predicate is mapped into a set of constraints over point coordinates\footnote{For example, the predicate \textit{isPerpendicular}(AB, CD) is mapped to the constraint $\frac{y_B-y_A}{x_B-x_A} \times \frac{y_D-y_C}{x_D-x_C} = -1$.}. These constraints can be non-trivial to write, requiring significant manual engineering. As a result, \textit{GEOS}'s constraint set is incomplete and it cannot solve a number of SAT style geometry questions. Furthermore, this solver is not interpretable. As our user studies show, it is not natural for a student to understand the solution of these geometry questions in terms of satisfiability of constraints over coordinates. A more natural way for students to understand and reason about these questions is through deductive reasoning using axioms of geometry.
% Optimizing the sum of indicator functions is a difficult (non-convex) optimization problem and hence relaxations are introduced to solve it. This leads to stability issues in the solver. Finally, as this is a numerical solver, it has precision issues and the answer is usually only an approximation of the correct answer.

\section{Set up for the Axiomatic Solver}\label{sec:solver}
\begin{figure}
	%\begin{minipage}[c]{0.627\textwidth}
	\includegraphics[width=0.75\textwidth]{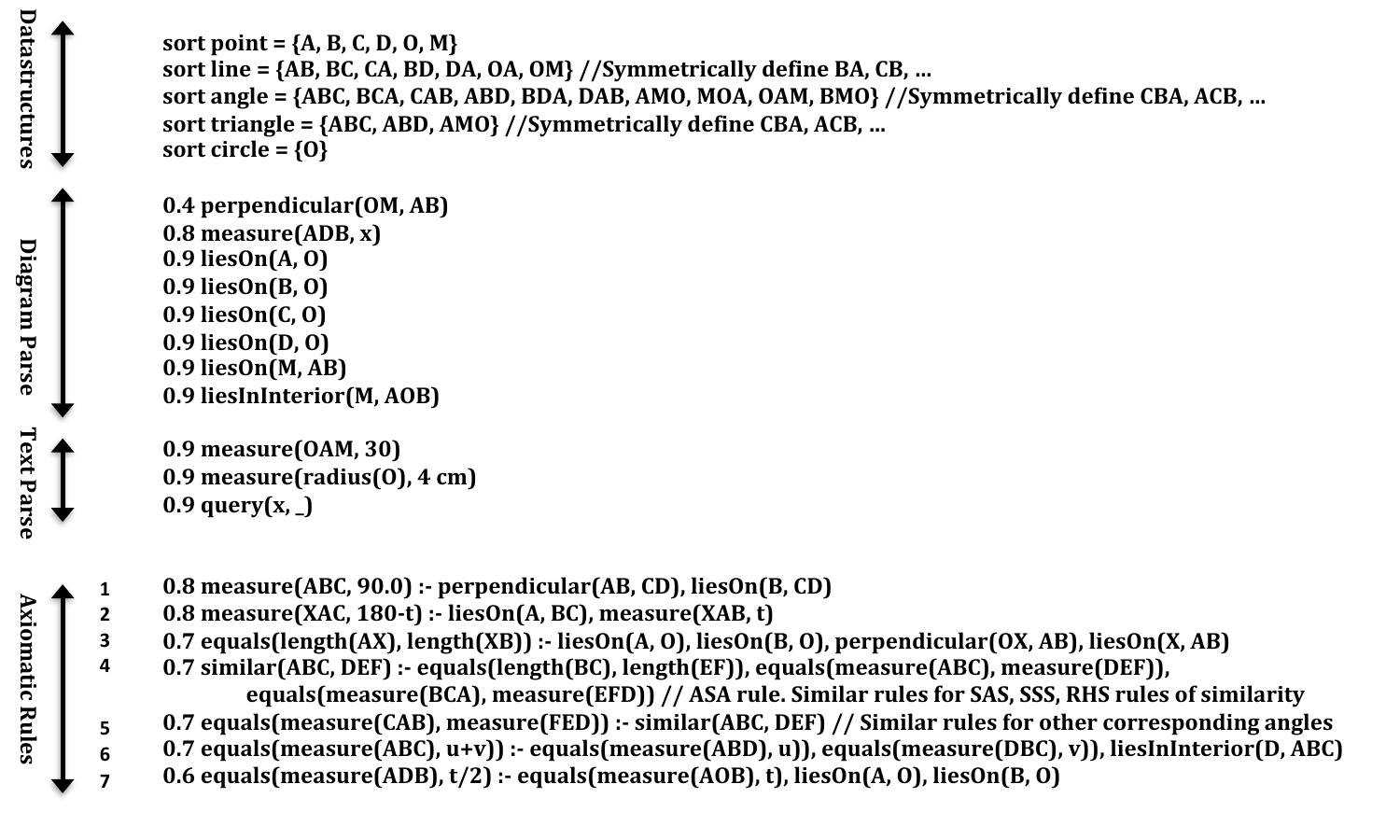}
	%\end{minipage}\hfill
	%\begin{minipage}[c]{0.373\textwidth}
	\caption{A sample logical program (in prolog style) that solves the problem in Figure \ref{fig:geo-que}. The program consists of a set of data structure declarations that correspond to types in the prolog program, a set of declarations from the diagram and text parse and a subset of the geometry axioms written as horn clause rules. The axioms are used as the underlying theory with the aforementioned declarations to yield the solution upon logical inference. Normalized confidence weights from the diagram, text and axiom parses are used as probabilities. For readers understanding, we list the axioms in the order (1 to 7) they are used to solve the problem. However, this ordering is not required. Other (less probable) declarations and axiom rules are not shown here for clarity but they can be assumed to be present.}\label{fig:problog}
	%\end{minipage}
\end{figure}
To tackle the aforementioned issues with the numerical solver in \textit{GEOS}, we replace the numerical solver with an axiomatic solver. We extract axiomatic knowledge from textbooks and parse them into horn clause rules. Then we build an axiomatic solver that performs logical inference with these horn clause rules and the formal problem description.
A sample logical program (in prolog notation) that solves the problem in Figure \ref{fig:example} is given in Figure \ref{fig:problog}. The logical program has a set of declarations from the \textit{GEOS} text and diagram parsers which describe the problem specification and the parsed horn clause rules describe the underlying theory. Normalized confidence scores from question text, diagram and axiom parsing models are used as probabilities in the program.
Next, we describe how we harvest structured axiomatic knowledge from textbooks.

\section{Harvesting Axiomatic Knowledge}\label{sec:thm-parse}
%Textbooks are rich sources of semi-structured knowledge such as axioms. 
We present a structured prediction model that identifies axioms in textbooks and then parses them. Since harvesting axioms from a single textbook is a very hard problem, we use multiple textbooks and leverage the redundancy of information to accurately extract and parse axioms. We first define a joint model that identifies axiom mentions in each textbook and aligns repeated mentions of the same axiom across textbooks. Then, given a set of axioms (with possibly, multiple mentions of each axiom), we define a parsing model that maps each axiom to a horn clause rule by utilizing the various mentions of the axiom.

Given a set of textbooks $\mathcal{B}$ in machine readable form (json in our experiments), we extract chapters relevant for geometry in each of them to obtain a sequence of discourse elements (with associated typographical information) from each textbook.
We assume that the textbook comprises of an ordered set\footnote{Given a textbook in json format, we can construct this ordered set by preorder traversal of the json tree.} of \textit{discourse elements} where a discourse element could be a natural language sentence, heading, title, figure, table or a caption. The discourse element (for example, a sentence) could have additional typographical features. For example, the sentence could be written in boldface, underline, etc. These properties of discourse elements will be useful features which can be leveraged for the task of harvesting axioms.
Let ${\bf S}_b = \{s_0^{(b)}, s_1^{(b)}, \dots s_{|{\bf S}_b|}^{(b)}\}$ denote the sequence of discourse elements in textbook $b$. $|{\bf S}_b|$ denotes the number of discourse elements in textbook $b$.

\subsection{Axiom Identification and Alignment}
%We first describe our models for axiom identification and alignment.
We decompose the problem of extracting axioms from textbooks into two tractable sub-problems:
\begin{enumerate}
	\item identification of axiom mentions in each textbook using sequence labeling
	\item aligning repeated mentions of the same axiom across textbooks
\end{enumerate}
Then, we combine the learned models for these sub-problems into a joint optimization framework that simultaneously learns to identify and align axiom mentions. Joint modeling of the axiom identification and alignment is necessary as both sub-problems can help each other.
%\input{GM/cim-illustration}
%Figure \ref{fig:CIM} shows an instantiation of this model.
% While better axiom identification would lead to better axiom alignment for obvious reasons, axiom alignment can help in axiom identification as the alignment information can be used to design features that match an axiom mention with axiom mentions aligned to it.

\subsubsection{Axiom Identification}
Linear-chain CRF formulation \cite{lafferty:2001} can be used for the subproblem of axiom identification. Given $\{{\bf S}_b|b \in \mathcal{B}\}$, a sequence of discourse elements (with associated typographical information) from each textbook, 
the model labels each discourse element $s_i^{(b)}$ as {\bf B}efore, {\bf I}nside or {\bf O}utside an axiom.
Hereon, a contiguous block of discourse elements labeled ${\bf B}$ or ${\bf I}$ will be considered as an axiom mention.
Let $\mathcal{T} = \{{\bf B}, {\bf I}, {\bf O}\}$ denote the tag set.  Let $y_i^{(b)}$ be the tag assigned to $s_i^{(b)}$ and ${\bf Y}_{b}$ be the tag sequence assigned to ${\bf S}_b$. The CRF defines:

\begin{equation*}
p({\bf Y}_b | {\bf S}_b;\pmb{\theta}) \propto \prod\limits_{k=1}^{|{\bf S}_b|}\exp\left( \sum\limits_{i, j\in \mathcal{T}}\pmb{\theta}_{ij}^T{\bf f}_{ij}(y_{k-1}^{(b)}, y_k^{(b)}, \mathcal{S}_b) \right)
\end{equation*}
%Here, ${\bf f}_{ij}$ are feature functions defined later. 
We find the parameters $\pmb{\theta}$ using maximum-likelihood estimation with L2 regularization:

\begin{equation*}
\pmb{\theta}^* = \argmax_{\pmb{\theta}} \sum\limits_{b \in \mathcal{B}} \log p({\bf Y}_b | {\bf S}_b;\pmb{\theta}) - \lambda||\pmb{\theta}||_2^2
\end{equation*}
We use L-BFGS to optimize the objective and Viterbi decoding for inference. $\lambda$ is tuned on the dev set.
\\
{\bf Features:} Features $f$ look at a pair of adjacent tags $y_{k-1}^{(b)}$, $y_k^{(b)}$, the input sequence ${\bf S}_b$, and where we are in the sequence. The features (listed in Table \ref{tab:feats1}) include various content based features encoding various notions of similarity between pairs of discourse elements (in terms of semantic overlap, more refined match of geometry entities and certain keywords) as well as various typographical features such as whether the discourse elements are annotated as an axiom (or theorem or corollary) in the textbook, contain equations, diagrams, text that is bold or italicized, are in the same node of the json hierarchy, are contained in a bounding box, etc. We also use features difectly from an existing RST parser \cite{feng-hirst:2014} -- discourse structure can be useful to understand if two consecutive discourse elements are together part of an axiom (or not).

Some extracted axiom mentions contain pointers to a diagram eg. ``Figure 2.1''. In all these cases, we consider the diagram to be a part of the axiom mention. We will discuss the impact of the various content and typpgraphy based features later in section \ref{sec:feats}.
\begin{table*}
	\center
	\begin{tabular}{|c|p{1.5cm}|p{10.3cm}|}
		\hline
		\parbox[t]{1mm}{\multirow{12}{*}{\rotatebox[origin=c]{90}{Content}}} & Sentence Overlap & Semantic Textual Similarity between the current and next discourse element. We include features that compute the proportion of common unigrams and bigrams across the two discourse elements. This feature is conjoined with the tag assigned to the current and next sentence.\\\cline{2-3}
		& Geometry entities & No. of geometry entities (constants, predicates and functions) -- normalized by the number of tokens in this discourse element. This feature is conjoined with the tag assigned to the current discourse element.\\\cline{2-3}
		& Keywords & Indicator that the current discourse element contains any one of the following words: \textit{hence}, \textit{if}, \textit{equal}, \textit{twice}, \textit{proportion}, \textit{ratio}, \textit{product}. This feature is conjoined with the tag assigned to the current discourse element. \\\hline\hline
		
		\parbox[t]{1mm}{\multirow{22}{*}{\rotatebox[origin=c]{90}{Discourse}}} 		& RST edge &  Indicator for the RST relation between the current and next discourse element. This feature is conjoined with the tag assigned to the current and next sentence.\\\cline{2-3}
		& Axiom, Theorem, Corollary Mention & (a) The current (or previous) discourse element is mentioned as an Axiom, Theorem or Corollary e.g. \textit{Similar Triangle Theorem} or \textit{Corollary 2.1}. \newline (b) The section or subsection in the textbook containing the current (or previous) discourse element mentions an Axiom, Theorem or Corollary. \newline This feature is conjoined with the tag assigned to the current (and previous) discourse element.\\\cline{2-3}
		& Equation & The current (or next) discourse element contains an equation eg. $PA \times PB = PT^2$. This feature is conjoined with the tag assigned to the current (and next) sentence.\\\cline{2-3}
		& Associated Diagram & The current discourse element contains a pointer to a figure eg. ``Figure 2.1''. This feature is conjoined with the tag assigned to the current discourse element.\\\cline{2-3}
		& Bold / Underline & The discourse element (or previous discourse element) contains text that is in bold font or underlined. Conjoined with the tag assigned to the current (and previous) discourse element.\\\cline{2-3}
		& Bounding box &  Indicator that the current and previous discourse elements are bounded by a bounding box in the textbook. Conjoined with the tag assigned to the current and previous discourse element.\\	\cline{2-3}
		& JSON structure &  Indicator that the current and previous discourse element are in the same node of the JSON hierarchy. Conjoined with the tag assigned to the current and previous discourse elements.\\\hline
	\end{tabular}
	\caption{Feature set for our axiom identification model. The features are based on content and typography.\label{tab:feats1}}
\end{table*}

\subsubsection{Axiom Alignment}\label{sampler}
Next, we leverage 
%two keys sources of information for harvesting axiomatic knowledge: 
the redundancy of information and the relatively fixed ordering of axioms in various textbooks. Most textbooks typically present all axioms of geometry in approximately same order, moving from easier concepts to more advanced concepts. For example, all textbooks will introduce the definition of a right angled triangle before introducing the Pythagorous theorem. We leverage this structure by aligning various mentions of the same axiom across textbooks and introducing structural constraints on the alignment.
%and alignment.

Let ${\bf A}_b = \left(A_1^{(b)}, A_2^{(b)}, \dots, A_{|{\bf A}_b|}^{(b)}\right)$ be the axiom mentions extracted from textbook $b$. Let ${\bf A}$ denote the collection of axiom mentions extracted from all textbooks.
%Let $Z_{ij}^{(b_1, b_2)}$ denote the random variable which denotes if $A_i^{(b_1)}$ is aligned to $A_j^{(b_2)}$.  Let ${\bf Z}$ denote the set of all alignments. 
%We introduce a log-linear model that factorizes over alignment pairs:
We assume a global ordering of axioms ${\bf A}^* = \left(A_1^*, A_2^*, \dots, A_U^*\right)$ where $U$ is some pre-defined upper bound on the total number of axioms in geometry. Then, we emphasize that the axiom mentions extracted from each textbooks (roughly) follow this ordering. Let $Z_{ij}^{(b)}$ be a random variable that denotes if axiom $A_i^{(b)}$ extracted from book $b$ refers to the global axiom $A_j^*$.
%Let ${\bf Z}$
%= \bigcup\limits_{\substack{b\in\mathcal{B}\\1\leq i \leq |{\bf A}_b|\\\\1\leq j \leq U}} Z_{ij}^{(b)}$ 
%be the collection of all these indicator variables. 
%Using the indicator variables defined above
We introduce a log-linear model that factorizes over alignment pairs:
\begin{equation*}
P({\bf Z}| {\bf A};\pmb{\phi}) = \frac{1}{Z({\bf A};\pmb{\phi})} \times
\exp\left( \sum\limits_{\substack{b_1, b_2 \in \mathcal{B}\\b_1 \neq b_2}}\sum\limits_{1 \leq k \leq U}\sum\limits_{\substack{1 \leq i \leq |{\bf A}_{b_1}|\\1 \leq j \leq |{\bf A}_{b_2}|}} Z_{ik}^{(b_1)} Z_{jk}^{(b_2)} \pmb{\phi}^T{\bf g}(A_i^{(b_1)}, A_j^{(b_2)}) \right)\nonumber
\end{equation*}

Here, $Z({\bf A};\pmb{\phi})$ is the partition function of the log-linear model.
%$\sigma(x) = \frac{1}{1+e^{-x}}$ denotes the sigmoid function and 
${\bf g}$ denotes a feature function that measures the similarity of two axiom mentions (described in detail later).
%Next, using these random variables, w
We introduce the following constraints on the alignment structure:
\\
{\bf C1:} An axiom appears in a book at-most once
%	$\sum_i Z_{ij}^{(b)} \leq 1 \hspace{0.5cm} \forall \hspace{0.2cm} 1 \leq j \leq U \hspace{0.5cm} \forall b \in \mathcal{B}$
\\
{\bf C2:} An axiom refers to exactly one theorem in the global ordering
%:\\
%$\sum_j Z_{ij}^{(b)} = 1 \hspace{0.5cm} \forall \hspace{0.2cm} 1 \leq i \leq |A_b| \hspace{0.5cm} \forall b \in \mathcal{B}$
\\
{\bf C3:} Ordering Constraint: If $i^{th}$ axiom in a book refers to the $j^{th}$ axiom in the global ordering then no axiom succeeding the $i^{th}$ axiom can refer to a global axiom preceding $j$.\\
%:\\
%$\hspace*{0.1cm}Z_{ij}^{(b)} \leq 1 - Z_{kl}^{(b)}$ \\
%$\hspace*{1cm} \forall\hspace{0.2cm} 1 \leq i < k \leq |{\bf A}_b|, 1 \leq l < j \leq U$\\
%$\hspace*{1cm} \forall\hspace{0.2cm} b \in \mathcal{B}$
\\
{\bf Learning with Hard Constraints:}
We find the optimal parameters $\pmb{\phi}$ using maximum-likelihood estimation with L2 regularization:

\begin{equation*}
\pmb{\phi}^* = \argmax_{\pmb{\phi}} \log P({\bf Z} | {\bf A};\pmb{\phi}) - \mu||\pmb{\phi}||_2^2
\end{equation*}
We use L-BFGS to optimize the objective.
% The gradient of the above objective is: 
%\\
%{\scriptsize
%$
%	\sum\limits_{\substack{b_1, b_2 \in \mathcal{B}\\b_1 \neq b_2}}\sum\limits_{1 \leq k \leq U}\sum\limits_{\substack{1 \leq i \leq |{\bf A}_{b_1}|\\1 \leq j \leq |{\bf A}_{b_2}|}} Z_{ik}^{(b_1)} Z_{jk}^{(b_2)} {\bf g}(A_i^{(b_1)}, A_j^{(b_2)}) -\hspace{4.5cm}\\
%	\mathbb{E}_{P({\bf Z}| {\bf A};\pmb{\phi})} \Big[ \sum\limits_{\substack{b_1, b_2 \in \mathcal{B}\\b_1 \neq b_2}}\sum\limits_{1 \leq k \leq U}\sum\limits_{\substack{1 \leq i \leq |{\bf A}_{b_1}|\\1 \leq j \leq |{\bf A}_{b_2}|}} Z_{ik}^{(b_1)} Z_{jk}^{(b_2)} {\bf g}(A_i^{(b_1)}, A_j^{(b_2)}) \Big]
%$
%}
%\\
To compute feature expectations appearing in the gradient of the objective, we use a Gibbs sampler.
The sampling equations for $Z_{ik}^{b}$ are:
%\[p(Z_{ik}^{(b)}|rest) = \frac{\exp\left( T(k)\right)}{\sum\limits_{1 \leq k' \leq U}\exp\left( T(k') \right)}\]

\begin{eqnarray}
&&P(Z_{ik}^{(b)}|rest) \propto \exp\left( T_b(i, k) \right)\\\nonumber
&&T_b(i, k) = Z_{ik}^{(b)} \sum\limits_{\substack{b'\in \mathcal{B}\\b' \neq b}} \sum\limits_{1\leq j \leq |A_{b'}|} Z_{jk}^{(b')} \pmb{\phi}^T{\bf g}(A_i^{(b)}, A_j^{(b')})
\end{eqnarray}

Note that the constraints $C1\dots 3$ define the feasible space of alignments. Our sampler always samples the next $Z_{ik}^{(b)}$ in this feasible space.
$\mu$ is tuned on the development set.
\\
{\bf Learning with Soft Constraints:} We might want to treat some constraints, in particular, the ordering constraints $C3$ as soft constraints. We can write down the constraint $C3$ using the alignment variables:\\
$\hspace*{4.1cm}Z_{ij}^{(b)} \leq 1 - Z_{kl}^{(b)}$ \\
$\hspace*{5cm} \forall\hspace{0.2cm} 1 \leq i < k \leq |{\bf A}_b|, 1 \leq l < j \leq U$\\
$\hspace*{5cm} \forall\hspace{0.2cm} b \in \mathcal{B}$

To model these constraints as soft constraints, we penalize the model for violating these constraints. Let the penalty for violating the above constraint be $\exp\left(\nu\max\left(0, 1 - Z_{ij}^{(b)} - Z_{kl}^{(b)} \right)\right)$.
%${\bf C3(Z)} \leq {\bf 0}$ where 
Thus, we introduce a new regularization term: 
\begin{equation*}
{\bf R(Z)} = \sum_{\substack{1 \leq i < k \leq |{\bf A}_b| \\ 1 \leq l < j \leq U \\ b \in \mathcal{B}}} \exp\left(\nu\max\left(0, 1 - Z_{ij}^{(b)} - Z_{kl}^{(b)} \right)\right)
\end{equation*}
Here $\nu$ is a hyper-parameter to tune the cost of violating a constraint.
% Then, the posterior can be written as:\\ 
%\begin{eqnarray}\label{eqn3}
%	p({\bf Z}| {\bf A}, C2;\pmb{\phi}) \propto
%	p({\bf Z}| {\bf A};\pmb{\phi}) \times \exp\left( \nu{\bf 1}^T{\bf C3}({\bf Z}) \right)\nonumber
%\end{eqnarray}\\
%We model the remaining constraints as hard constraints. 
We write down the following regularized objective:

\begin{equation*}
\pmb{\phi}^* = \argmax_{\pmb{\phi}} \log P({\bf Z} | {\bf A};\pmb{\phi}) - {\bf R(Z)} - \mu||\pmb{\phi}||_2^2
\end{equation*}

We use L-BFGS to find the optimal parameters $\pmb{\phi}*$.
We perform Gibbs sampling to compute feature expectations. The sampling equation for $Z_{ik}^{(b)}$ is similar (eq 1), but:
\begin{eqnarray*}
	T_b(i, k) = \sum\limits_{\substack{b'\in \mathcal{B}\\b' \neq b}} \sum\limits_{1\leq j \leq |A_{b'}|} Z_{ik}^{(b)}Z_{jk}^{(b')} \pmb{\phi}^T{\bf g}(A_i^{(b)}, A_j^{(b')})\\
	\hspace*{1.5cm}+\nu\sum\limits_{\substack{b'\in \mathcal{B}\\b' \neq b}}\sum\limits_{i < j \leq |A_{b'}|}\sum\limits_{1 \leq l < k}\left(1-Z_{ik}^{(b)}-Z_{jl}^{(b')}\right)\\
	\hspace*{1.8cm}+\nu\sum\limits_{\substack{b'\in \mathcal{B}\\b' \neq b}}\sum\limits_{1 \leq j < i|}\sum\limits_{k < l \leq U}\left(1-Z_{ik}^{(b)}-Z_{jl}^{(b')}\right)
\end{eqnarray*}
%\\
%{\bf Initialization:} The sampler starts with an initial assignment for the alignment variables obtained from another alignment model that trains a binary logistic regression classifier on the development set using axiom similarity features listed in Table \ref{tab:feats2} to classify whether a pair of axiom mentions is aligned or not and then enforces the alignment constraints using an ILP. Please refer to the supplementary for details. 
\\
{\bf Features:} Now, we describe the features $g$. These too include content based features encoding various notions of similarity between pairs of axiom mentions (such as unigram, bigram, dependency and entity overlap, longest common subsequence (LCS), alignment, MT and summarization scores) as well as various typographical features such matching  of the current (and parent) node of axiom mentions in respective json hierarchies, equation template matching and image caption matching. The features are listed in Table \ref{tab:feats2}. We will further discuss the impact of the various content and typpgraphy based features later in section \ref{sec:feats}.
\begin{table*}
	\center
	\begin{tabular}{|c|>{\raggedright}p{2.1cm}|p{10.4cm}|}
		\hline
		\parbox[t]{1mm}{\multirow{20}{*}{\rotatebox[origin=c]{90}{Content}}} & Unigram, Bigram, Dependency and Entity Overlap & Real valued features that compute the proportion of common unigrams, bigrams, dependencies and geometry entities (constants, predicates and functions) across the two axioms. When comparing geometric entities, we include geometric entities derived from the associated diagrams when available.\\\cline{2-3}
		&Longest Common Subsequence & Real valued feature that computes the length of longest common sub-sequence of words between two axiom mentions normalized by the total number of words in the two mentions.\\\cline{2-3}
		&Number of discourse elements & Real valued feature that computes the absolute difference in the number of discourse elements in the two mentions.\\\cline{2-3}
		&Alignment Scores & We use an off-the-shelf monolingual word aligner -- \textit{JACANA} \cite{Yaox:2013} pretrained on PPDB -- and compute alignment score between axiom mentions as the feature.\\\cline{2-3}
		&MT Metrics & We use two common MT evaluation metrics \textit{METEOR} \cite{denkowski:2010} and \textit{MAXSIM} \cite{chan:2008}, and use the evaluation scores as features. While \textit{METEOR} computes n-gram overlaps controlling on precision and recall, \textit{MAXSIM} performs bipartite graph matching and maps each word in one axiom to at most one word in the other.\\\cline{2-3}
		&Summarization Metrics & We also use \textit{Rouge-S} \cite{lin:2004}, a text summarization metric, and use the evaluation score as a feature. \textit{Rouge-S} is based on skip-grams.\\\hline
		\parbox[t]{1mm}{\multirow{11}{*}{\rotatebox[origin=c]{90}{Discourse (Typography)}}} &JSON structure &  Indicator matching the current (and parent) node of axiom mentions in respective JSON hierarchies, i.e. are both nodes mentioned as axioms, diagrams or bounding boxes?\\\cline{2-3}
		&Equation Template & Indicator feature that matches templates of equations detected in the axiom mentions. The template matcher is designed such that it identifies various rewritings of the same axiom equation e.g. $PA \times PB = PT^2$ and $PA \times PB = PC^2$ could refer to the same axiom with point $T$ in one axiom mention being point $C$ in another mention.\\\cline{2-3}
		&Image Caption & Proportion of common unigrams in the image captions of the diagrams associated with the axiom mentions. If both mentions do not have associated diagrams, this feature doesn't fire.\\\hline
	\end{tabular}
	\caption{Feature set for our axiom alignment model. The features are based on content and typography.\label{tab:feats2}}
\end{table*}

\subsubsection{Joint Identification and Alignment}
%As described before, j

Joint modeling of axiom identification and alignment components is useful as both problems potentially help each other.
Hence, we combine the respective models for identification and alignment into a joint model.
%In this case, w
Let $Y_{ij}^{(b)}$ denote that the discourse element $s_i^{(b)}$ from book $b$ has tag $j$.
%, and 
We reuse the definitions of the alignment variables $Z_{ij}^{(b)}$ as before.
%which denotes that the sentence $s_i^{(b)}$ is part of an axiom referring to the global axiom $A_j^*$. 
We further define $Z_{i0}^{(b)}$ such that it denotes that the $i^{th}$ axiom in textbook $b$ is not aligned to any global axiom. 
We again define a log-linear model with factors that score axiom identification and axiom alignments.
\begin{equation*}
p({\bf Y}, {\bf Z}| \{{\bf S}_b\};\pmb{\theta}, \pmb{\phi}) \propto f_{AI}({\bf Y}| \{{\bf S}_b\};\pmb{\theta}) \times f_{AA}({\bf Z}| {\bf Y}, \{{\bf S}_b\};\pmb{\phi})
\end{equation*}
Here, the factors:
\begin{eqnarray*}
	f_{AI} = \exp(\sum\limits_{b \in \mathcal{B}} \sum\limits_{k=1}^{|{\bf S}_b|} \sum\limits_{i, j\in \mathcal{T}}Y_{k-1i}^{(b)}Y_{kj}^{(b)}\pmb{\theta}_{ij}^T{\bf f}_{ij}(i, j, \mathcal{S}_b))\hspace{2.1cm}\\
	f_{AA} = \exp(\sum\limits_{\substack{b_1, b_2 \in \mathcal{B}\\b_1 \neq b_2}}\sum\limits_{1 \leq k \leq U}\sum\limits_{\substack{1 \leq i \leq |{\bf A}_{b_1}|\\1 \leq j \leq |{\bf A}_{b_2}|}} Z_{ik}^{(b_1)} Z_{jk}^{(b_2)} \pmb{\phi}^T{\bf g}(A_i^{(b_1)}, A_j^{(b_2)}))
\end{eqnarray*}
%{\small 
%	\begin{eqnarray}\label{eqn4}
%	p({\bf Y}, {\bf Z}| \{{\bf S}_b\};\pmb{\theta}, \pmb{\phi}) \propto \hspace{5.35cm}\nonumber\\
%	\exp(\sum\limits_{b \in \mathcal{B}} \sum\limits_{k=1}^{|{\bf S}_b|} \sum\limits_{i, j\in \mathcal{T}}Y_{k-1i}^{(b)}Y_{kj}^{(b)}\pmb{\theta}_{ij}^T{\bf f}_{ij}(i, j, \mathcal{S}_b) +\hspace{2cm}\nonumber\\
%	\sum\limits_{\substack{b_1, b_2 \in \mathcal{B}\\b_1 \neq b_2}}\sum\limits_{1 \leq k \leq U}\sum\limits_{\substack{i: Y_{iB}^{(b_1)} = 1\\j: Y_{jB}^{(b_2)} = 1}} Z_{ik}^{(b_1)} Z_{jk}^{(b_2)} \pmb{\phi}^T{\bf g}(A_i^{(b_1)}, A_j^{(b_2)}))\nonumber
%	\end{eqnarray}
%}
We write down the model constraints below:
\\
{\bf C1'}: Every discourse element has a unique label
%:\\
%$\sum\limits_{j} Y_{ij}^{(b)} = 1 \hspace{0.5cm} \forall \hspace{0.2cm} b \in \mathcal{B}, 1 \leq i \leq |{\bf S}_b|$
\\
{\bf C2'} Tag O cannot be followed by tag I
%:\\
%$Y_{i-1O}^{(b)} \leq 1-Y_{i+1I}^{(b)} \hspace{0.5cm} \forall \hspace{0.2cm} b \in \mathcal{B},  1 \leq i \leq |{\bf S}_b|$
\\
{\bf C3'} Consistency between $Y$'s and $Z$'s i.e. axiom boundaries defined by $Y$'s and $Z$'s must agree.
%: For a given axiom length $1 \leq l \leq M$ for some upper limit $M$, the sentence-tag labelling $Y$ agrees with the sentence to axiom labeling $Z$, i.e.,  and($Y_{iB}, Y_{i+1I}, Y_{i+2I}, \dots, Y_{i+kI}, !Y_{i+k+1I}$) $\iff$ and($Z_{it}, Z_{i+1t}, Z_{i+2t}, \dots, Z_{i+kt}, Z_{i+k+1t'}$):\\
%$Y^{(and1)}_{ik} = Y^{(and2)}_{iktt'}$\\
%$Y^{(and1)}_{ik} \leq Y_{iB}$\\
%$Y^{(and1)}_{ik} \leq Y_{i+jI} \hspace{1cm} \forall 1 \leq j \leq k$\\
%$Y^{(and1)}_{ik} \leq 1-Y_{i+k+1I}$\\
%$Y^{(and1)}_{ik} \geq Y_{iB} + Y_{i+1I} + Y_{i+2I}, \dots + Y_{i+kI} + 1-Y_{i+k+1I}$\\
%$Y^{(and2)}_{iktt'} \geq $
%We express the boolean logic operations (and, not, iff) in zero-one integer-linear-programming\footnote{http://cs.stackexchange.com/questions/12102/express-boolean-logic-operations-in-zero-one-integer-linear-programming-ilp}.
\\
{\bf C4'} = C3.
%We sample $Y$ and $Z$ alternatively.
%The sampling equations are:
%\[p(Y_{ij}^{(b)}, Z_{ik}^{(b)}|rest) = \frac{\exp\left( T(j, k)\right)}{\sum\limits_{\substack{j' \in \{B, I, O\}\\1 \leq k' \leq U}}\exp\left( T(j', k') \right)}\]
%where, \\$T(j, k) = \sum\limits_{u \in \{B, I, O\}}Y_{i-1u}^{(b)}Y_{ij}^{(b)} \pmb{\theta}_{iu}^T{\bf f}(u, j, S_b) +\\\hspace*{2cm}
%\sum\limits_{\substack{b'\in \mathcal{B}\\b' \neq b}} \sum\limits_{1\leq j \leq |A_{b'}|} Z_{ik}^{(b)}Z_{jk}^{(b')} \pmb{\phi}^T{\bf g}(A_i^{(b)}, A_j^{(b')}) +\\\hspace*{2cm} \lambda\sum\limits_{\substack{b'\in \mathcal{B}\\b' \neq b}}\sum\limits_{i < j \leq |A_{b'}|}\sum\limits_{1 \leq l < k}\left(1-Z_{ik}^{(b)}-Z_{jl}^{(b')}\right) +\\\hspace*{2.5cm} \lambda\sum\limits_{\substack{b'\in \mathcal{B}\\b' \neq b}}\sum\limits_{1 \leq j < i|}\sum\limits_{k < l \leq U}\left(1-Z_{ik}^{(b)}-Z_{jl}^{(b')}\right)$
%Every time we obtain a sample for $Y$ that changes axiom boundaries, we sample the alignment variables $Z$ along with it.
%To model the constraints, w

\begin{figure}
	\center
	\includegraphics[scale=0.312]{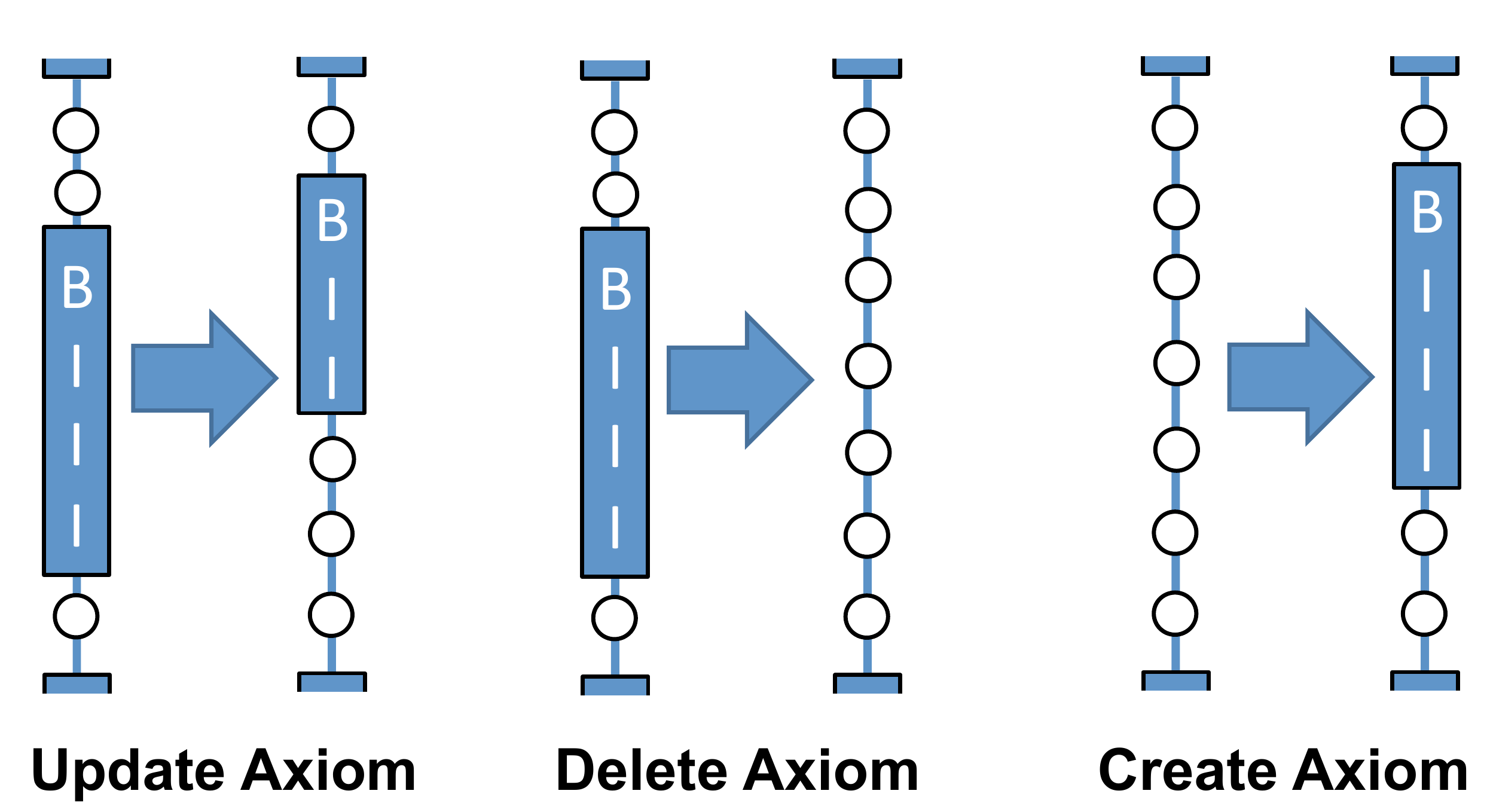}
	\caption{An illustration of the three operations to sample axiom blocks.}\label{fig:gibbs}
\end{figure}
We use L-BFGS for learning. To compute feature expectations, we use a Metropolis Hastings sampler that samples ${\bf Y}'s$ and ${\bf Z}'s$ alternatively. Sampling for ${\bf Z}'s$ reduces to Gibbs sampling and the sampling equations are as same as before (Section \ref{sampler}). For better mixing, we sample ${\bf Y}$ in blocks. Consider blocks of ${\bf Y}$'s which denote axiom boundaries at time stamp $t$ , we define three operations to sample axiom blocks at the next time stamp. The operations (shown in Figure \ref{fig:gibbs}) are:
\\
{\bf Update axiom:} The axiom boundary can be shrunk, expanded or moved. The new axiom, however, cannot overlap with other axioms.
\\
{\bf Delete axiom:} The axiom can be deleted by labeling all its discourse elements as $O$.
\\
{\bf Introduce axiom:} Given a contiguous sequence of discourse elements labeled $O$, a new axiom can be introduced.
\\
Note that these three operations define an ergodic Markov chain. We use the axiom identification part of the model as the proposal: 

\begin{equation*}
Q({\bf \bar Y}|{\bf Y}) \propto \exp\left(\sum\limits_{b \in \mathcal{B}} \sum\limits_{k=1}^{|{\bf S}_b|} \sum\limits_{i, j\in \mathcal{T}}\bar Y_{k-1i}^{(b)}\bar Y_{kj}^{(b)}\pmb{\theta}_{ij}^T{\bf f}_{ij}(i, j, \mathcal{S}_b)\right)
\end{equation*}
Hence, the acceptance ratio only depends on the alignment part of the model: $R({\bf \bar Y}|{\bf Y}) = \min\left(1, \frac{U({\bf \bar Y})}{U({\bf Y})}\right)$ where $U({\bf Y}) = f_{AA}$.
%\[p(Y_{ij}^{(b)}, Z_{ik}^{(b)}|rest) = \frac{\exp\left( T(j, k)\right)}{\sum\limits_{\substack{j' \in \{B, I, O\}\\1 \leq k' \leq U}}\exp\left( T(j', k') \right)}\]
%where, \\$T(j, k) = \sum\limits_{u \in \{B, I, O\}}Y_{i-1u}^{(b)}Y_{ij}^{(b)} \pmb{\theta}_{iu}^T{\bf f}(u, j, S_b) +\\\hspace*{2cm}
%\sum\limits_{\substack{b'\in \mathcal{B}\\b' \neq b}} \sum\limits_{1\leq j \leq |A_{b'}|} Z_{ik}^{(b)}Z_{jk}^{(b')} \pmb{\phi}^T{\bf g}(A_i^{(b)}, A_j^{(b')}) +\\\hspace*{2cm} \lambda\sum\limits_{\substack{b'\in \mathcal{B}\\b' \neq b}}\sum\limits_{i < j \leq |A_{b'}|}\sum\limits_{1 \leq l < k}\left(1-Z_{ik}^{(b)}-Z_{jl}^{(b')}\right) +\\\hspace*{2.5cm} \lambda\sum\limits_{\substack{b'\in \mathcal{B}\\b' \neq b}}\sum\limits_{1 \leq j < i|}\sum\limits_{k < l \leq U}\left(1-Z_{ik}^{(b)}-Z_{jl}^{(b')}\right)$
%Every time we obtain a sample for $Y$ that changes axiom boundaries, we sample the alignment variables $Z$ along with it.
%\subsection{Timing and Efficiency}
We again have two variants, where we model the ordering constraints (C$4'$) as soft or hard constraints.

\subsection{Axiom Parsing}
After harvesting axioms, we build a parser for these axioms that maps raw axioms to horn clause rules.
The axiom harvesting step provides us a multi-set of axiom extractions. Let $\mathcal{A} = \{{\bf A}_1, {\bf A}_2, \dots, {\bf A}_{|\mathcal{A}|}\}$ represent the multi-set where each axiom ${\bf A}_i$ is mentioned at least once.
Each axiom mention, in turn, comprises of a contiguous sequence of discourse elements and optionally an accompanying diagram.

Semantic parsers map natural language to formal programs such as database queries \cite[inter alia]{liang:2011}, commands to robots \cite[inter alia]{Shimizu:2009}, or even general purpose programs \cite{yin:17}. More specifically, \cite{Chang:2016} learn ``If-Then'' program statements and \cite{Quirk:15} learn ``If-This-Then-That'' rules. In theory, these works can be used to parse axioms to horn-clause rules.
However, semantic parsing is a hard task and would require a large amount of supervision. In our setting, we can only afford a modest amount of supervision. We mitigate this issue by using the redundant axiom mention extractions from multiple sources (textbooks) and combining the parses obtained from various textbooks to achieve a better final parse for each axiom.

First, we describe a base parser that parses axiom mentions to horn clause rules.
Then, we utilize the redundancy of axiom extractions from various sources (textbooks) to improve our parser.

\subsubsection{Base Axiomatic Parser}
%Before describing our multi-source axiomatic parser, we describe a base parser that parses axiom mentions to horn clause rules.
%Geometry axioms always have a premise and a conclusion.
%Hence, o
Our base parser identifies the \textit{premise} and \textit{conclusion} portions of each axiom and then uses \textit{GEOS}'s text parser to parse the two portions into a logical formula. Then, the two logical formulas are put together to form horn clause rules.

Axiom mentions (for example, the Pythagoras theorem mention in Figure \ref{fig:pythagorous}) are often accompanied by equations or diagrams. When the mention has an equation, we simply treat the equation as the \textit{conclusion} and the rest of the mention as the \textit{premise}. When the axiom has an associated diagram, we always include the diagram in the \textit{premise}. We learn a model to predict the split of the axiom text into two parts forming the \textit{premise} and the \textit{conclusion} spans. Then, the \textit{GEOS} parser maps the \textit{premise} and \textit{conclusion} spans to \textit{premise} and \textit{conclusion} logical formulas, respectively.
%using a log-linear model.

Let $Z_{s}$ represent the split that demarcates the \textit{premise} and \textit{conclusion} spans.
%Let $Z_{p}$ and $Z_{c}$ be the logical formulas for premise and conclusion respectively.
We score the axiom split
%, the premise and conclusion parses jointly 
as a log-linear model:
$
p(Z_{s}| a; {\bf w}) \propto 
\exp\left( {\bf w}^T{\bf h}(a, Z_{s}) \right)
$.
Here, ${\bf h}$ are feature functions described later.
%, ${\bf f_2}$ are feature functions to help parse the premise/conclusion. 
%The logical formulas $Z_{p}$ and $Z_{c}$ are obtained using a two part approach similar to \textit{GEOS}. It maps words or phrases in the text to corresponding concepts in the geometry language. Then, it identifies relations between identified concepts. This is achieved by decomposing the model ${\bf w_2}^T{\bf f}_2(a, Z)$ into two parts i.e. ${\bf w_2}^T{\bf f}_2(a, Z) = {\bf w_{21}}^T{\bf f}_{21}(a, Z) + {\bf w_{22}}^T{\bf f}_{22}(a, Z)$.
We found that in most cases ($>$95\%), the premise and conclusion are contiguous spans in the axiom mention where the left span corresponds to the \textit{premise} and the right span corresponds to the \textit{conclusion}. Hence, we search over the space of contiguous spans to infer $Z_{s}$.
Joint search over the latent variables: $Z_{s}$, $Z_{p}$ and $Z_{c}$ is exponential. Hence, we use a greedy procedure, beam search, with a fixed beam size (10) for inference. That is, in each step, we only expand the ten most promising candidates so far given by the current score.  We first infer $Z_{s}$ to decide the split of the axiom and then infer $Z_{p}$ and $Z_{c}$ to obtain the parse of the premise and the conclusion using the two-part approach described before. 
We use L-BGFGS for learning.\\
{\bf Features:} We list the features ${\bf h}$ defined over candidate spans forming the text split in Table \ref{tab:feats3}. 
The features are similar to those used in previous work on discourse analysis, in particular on the automatic detection of  \textit{elementary discourse units} (EDU's) in rhetorical structure theory \cite{mann:1988} and discourse parsing \cite{marcu:2000,soricut:2003}. %Features ${\bf f}_{2}$ are taken from \textit{GEOS}. 
These include ideas such as the use of a list of discourse markers, puncutations and natural text and json organization as an indicator of discourse boundaries. We also use an off-the-shelf discourse parser and an \textit{EDU} segmenter from \citet{soricut:2003}. Then we also used syntax based cues such as span lengths, head node attachment, distance to common ancestor/root, relative position of the two lexical heads and the text split, and \textit{dominance} which have been found to be useful in discourse parsing \cite{marcu:2000,soricut:2003}. Finally, we also used some semantic features such as the similarity of the two spans (in terms of common words, geometry relations and relation-arguments), number of geometry relations in the respective span parses. We will discuss the impact of the various features later in section \ref{sec:feats}.
% lists the features.
\begin{table*}
	\center
	\begin{tabular}{|c|>{\raggedright}p{1.7cm}|p{10.5cm}|}
		\hline
		\parbox[t]{1mm}{\multirow{12}{*}{\rotatebox[origin=c]{90}{Content}}} & Span Similarity & Proportion of (a) words, (b) geometry relations and (c) relation-arguments shared by the two spans.\\\cline{2-3}
		&No. of Relations & Number of geometry relations represented in the two spans. We use the Lexicon Map from \text{GEOS} to compute the number of expressed geometry relations.\\\cline{2-3}
		&Span Lengths & The distribution of the two text spans is typically dependent on their lengths. We use the ratio of the length of the two spans as an additional feature.\\\cline{2-3}
		&Relative Position & Relative position of the two lexical heads and the text split in the discourse element sentence. We use the difference between the lexical head position and the text split position as the feature.\\\hline
		\parbox[t]{1mm}{\multirow{36}{*}{\rotatebox[origin=c]{90}{Discourse (Typography)}}} & Discourse Markers & Discourse markers (connectives, cue-words or cue-phrases, etc) have been shown to give good indications on discourse structure \cite{marcu:2000}. We build a list of discourse markers using the training set, considering the first and last tokens of each span, culled to top 100 by frequency. We use these 100 discourse markers as features. We repeat the same procedure by using part-of-speech (POS) instead of words and use them as features.\\\cline{2-3}
		&Punctuation & Punctuation at the segment border is another excellent cue for the segmentation. We include indicator features whether there is a punctuation at the segment border.\\\cline{2-3}
		&Text Organization & Indicator that the two text spans are part of the same (a) sentence, (b) paragraph.\\\cline{2-3}
		&RST Parse & We use an off-the-shelf RST parser \cite{feng-hirst:2014} and include an indicator feature that the segmentation matches the parse segmentation. We also include the RST label as a feature.\\\cline{2-3}
		&Soricut and Marcu Segmenter & \citet{soricut:2003} (section 3.1) presented a statistical model for deciding elementary discourse unit boundaries. We use the probability given by this model retrained on our training set as feature. This feature uses both lexical and syntactic information.\\\cline{2-3}
		&Head / Common Ancestor/ Attachment Node & Head node is defined as the word with the highest occurrence as a lexical head in the lexicalized tree among all the words in the text span. The attachment node is the parent of the head node. We use features for the head words of the left and right spans, the common ancestor (if any), the attachment node and the conjunction of the two head node words. We repeat these features with part-of-speech (POS) instead of words.\\\cline{2-3}
		&Syntax & Distance to (a) root, and (b) common ancestor for the nodes spanning the respective spans. We use these distances, and the difference in the distances as features.\\\cline{2-3}
		&Dominance & \textit{Dominance} \cite{soricut:2003} is a key idea in discourse which looks at syntax trees and studies sub-trees for each span to infer a logical nesting order between the two. We use the dominance relationship as a feature. See \citet{soricut:2003} for details.\\\cline{2-3}
		&JSON Structure & Indicator that the two spans are in the same node in the JSON hierarchy. Conjoined with the indicator feature that the two spans are part of the same paragraph.\\\hline
	\end{tabular}
	\caption{Feature set for our axiom parsing model.\label{tab:feats3}}
\end{table*}
Given a beam of \textit{Premise} and \textit{Conclusion} splits, we use the \textit{GEOS} parser to get \textit{Premise} and \textit{Conclusion} logical formulas for each split in the beam and obtain a beam of axiom parses for each axiom in each textbook.
%\subsubsection{Post-processing to correct parses}

\subsubsection{Multi-source Axiomatic Parser}
Now, we describe a multi-source parser that utilizes the redundancy of axiom extractions from various sources (textbooks).  Given a beam of 10-best parses for each axiom from each source, we use a number of heuristics to determine the best parse for the axiom:
\\
{\bf 1. Majority Voting:} For each axiom, pick the parse that occurs most frequently across beams.\\
{\bf 2. Average Score:} Pick the parse that has the highest average parse score (only counting top 5 parses for each source), for each axiom.\\
{\bf 3. Learn Source Confidence:} Learn a set of weights $\{\mu_1, \mu_2, \dots, \mu_S\}$, one for each source and then picks the parse that has the highest average weighted parse score for each axiom.\\
{\bf 4. Predicate Score:} Instead of selecting from one of the top parses across various sources, treat each axiom parse as a bag of premise predicates and a bag of conclusion predicates. Then, pick a subset of premise and conclusion predicates for the final parse using average scoring with thresholding.

%\subsection{Joint Model for Axiom Identification and Parsing}
%\subsection{Learning from Weak, Strong and Mixed Supervision}\label{sec:learning}
%\subsection{Using Axiomatic Knowledge in the Logical Solver}
\section{Experiments}
{\bf Datasets and Baselines:} We use a collection of grade 6-10 Indian high school math textbooks by four publishers/authors -- \textit{NCERT}, \textit{R S Aggarwal}, \textit{R D Sharma} and \textit{M L Aggarwal} -- a total of $5 \times 4 = 20$ textbooks to validate our model. Millions of students in India study geometry from these books every year and these books are readily available online. We manually marked chapters relevant for geometry in these books and then parsed them using Adobe Acrobat's \textit{pdf2xml} parser and AllenAI's \textit{Science Parse} project\footnote{https://github.com/allenai/science-parse}. Then, we annotated geometry axioms, alignments and parses for grade 6, 7 and 8 textbooks by the four publishers/authors. We use grade 6, 7 and 8 textbook annotations for development, training, and testing, respectively. Grade 9 and 10 data are used as unlabeled data. Thus our method is semi-supervised. 
During training our axiom identification, alignment and joint axiom identification and alignment models, the latent variables ${\bf Z}$ are fixed for the training set and are not sampled. For the remaining data, these variables are sampled using our Gibbs sampler. All the hyper-parameters in all the models are tuned on the development set using grid search. Then, these hyper-parameter values are fixed and the entire training + development set is used for training (along with the unlabeled data) and all the models are evaluated on the test set.

\textit{GEOS} used 13 types of entities and 94 functions and predicates. We add some more entities, functions and predicates to cover other more complex concepts in geometry not covered in  \textit{GEOS}. Thus, we obtain a final set of 19 entity types and 115 functions and predicates for our parsing model. We use Stanford CoreNLP \cite{manning:2014} for feature generation.
We use two datasets for evaluating our system: (a) practice and official SAT style geometry questions used in \textit{GEOS}, and (b) an additional dataset of geometry questions collected from the aforementioned textbooks. This dataset consists of a total of 1406 SAT style questions across grades 6-10, and is approximately 7.5 times the size of the dataset used in \textit{GEOS}. We split the dataset into training (350 questions), development (150 questions) and test (906 questions) with equal proportion of grade 6-10 questions. We annotated the 500 training and development questions with ground-truth logical forms. We use the training set to train another version of \textit{GEOS} with expanded set of entity types, functions and predicates. We call this system \textit{GEOS++} which will be used as a baseline for our method.

\begin{table}
	%\begin{minipage}[c]{0.565\textwidth}
	\centering
	\begin{tabular}{c||c|c|c|c|c|c}
		& \multicolumn{3}{c|}{{\bf Strict Comp.}} & \multicolumn{3}{c}{{\bf Relaxed Comp.}}\\\cline{2-7}
		& {\bf P} & {\bf R} & {\bf F} & {\bf P} & {\bf R} & {\bf F} \tabularnewline
		\hline\hline
		{\bf Identification} & 64.3 & 69.3 & 66.7 & 84.3 & 87.9 & 86.1\\\hline
		{\bf Joint-Hard} & 68.0 & 68.1 & 68.0 & 85.4 & 87.1 & 86.2\\\hline
		{\bf Joint-Soft} & 69.7 & 71.1 & {\bf 70.4} & 86.9 & 88.4 & {\bf 87.6}\\
	\end{tabular}
	%\end{minipage}\hfill
	%\begin{minipage}[c]{0.425\textwidth}
	\caption{Test set Precision, Recall and F-measure scores for axiom identification when performed alone and when performed jointly with axiom alignment. We show results for both strict as well as relaxed comparison modes. For the joint model, we show results when we model ordering constraints as hard or soft constraints.\label{identify:res}}
	%\end{minipage}
\end{table}
\noindent{\bf Results:} We first evaluate the axiom identification, alignment and parsing models individually.

For axiom identification, we compare the results of automatic identification with gold axiom identifications and compute the precision, recall and F-measure on the test set. We use strict as well as relaxed comparison. In strict comparison mode the automatically identified mentions and gold mentions must match exactly to get credit, whereas, in the relaxed comparison mode only a majority ($>$50\%) of sentences in the automatically identified mentions and gold mentions must match to get credit. Table \ref{identify:res} shows the results of axiom identification where we clearly see improvements in performance when we jointly model axiom identification and alignment. This is due to the fact that both the components reinforce each other.
We also observe that modeling the ordering constraints as soft constraints leads to better performance than modeling them as hard constraints. This is because the ordering of presentation of axioms is generally (yet not always) consistent across textbooks.

\begin{table}
	\centering
	\begin{tabular}{c||c|c|c|c}
		& {\bf P} & {\bf R} & {\bf F} & {\bf NMI} \tabularnewline
		\hline\hline
		{\bf Alignment} & 71.8 & 74.8 & 73.3 & 0.60\\\hline
		{\bf Joint-Hard} & 75.0 & 76.4 & 75.7 & 0.65\\\hline
		{\bf Joint-Soft} & 79.3 & 81.4 & {\bf 80.3} & {\bf 0.69}\\
	\end{tabular}
	\caption{Test set Precision, Recall, F-measure and NMI scores for axiom alignment when performed alone and when performed jointly with axiom identification. For the joint model, we show results when we model ordering constraints as hard or soft constraints.\label{align:res}}
\end{table}
To evaluate axiom alignment, we first view it as a series of decisions, one for each pair of axiom mentions and compute precision, recall and F-score by comparing automatic decisions with gold decisions. Then, we also use a standard clustering metric, Normalized Mutual Information (NMI) \cite{strehl:2002} to measure the quality of axiom mention clustering.
Table \ref{align:res} shows the results on the test set when gold axiom identifications are used. We observe improvements in axiom alignment performance too when we jointly model axiom identification and alignment jointly both in terms of F-score as well as NMI.
Modeling ordering constraints as soft constraints again leads to better performance than modeling them as hard constraints in terms of both metrics.

\begin{table}
	%\begin{minipage}[c]{0.625\textwidth}
	\center
	\begin{tabular}{cc|c|c|c|c|c|c}
		& & \multicolumn{3}{c|}{{\bf Literals}} & \multicolumn{3}{c}{{\bf Full Parse}}\\\cline{3-8}
		& & {\bf P} & {\bf R} & {\bf F} & {\bf P} & {\bf R} & {\bf F} \tabularnewline
		\hline\hline
		&{\bf GEOS} & 86.7 & 70.9 & 78.0 & 64.2 & 56.6 & 60.2\\\hline
		\parbox[t]{0.3mm}{\multirow{5}{*}{\rotatebox[origin=c]{90}{{\bf GEOS++}}}}&\multicolumn{1}{|c|}{\begin{tabular}{@{}c@{}}{\bf Single Src.}\end{tabular}} & 91.6 & 75.3 & 82.6 & 68.8 & 60.4 & 64.3\\\cline{2-8}
		&\multicolumn{1}{|c|}{\begin{tabular}{@{}c@{}}{\bf Maj. Voting}\end{tabular}} & 90.2 & 78.5 & 83.9 & 70.0 & 63.3 & 66.5\\\cline{2-8}
		&\multicolumn{1}{|c|}{\begin{tabular}{@{}c@{}}{\bf Avg. Score}\end{tabular}} & 90.8 & 79.6 & 84.9 & 71.7 & 66.4 & 69.0\\\cline{2-8}
		&\multicolumn{1}{|c|}{\begin{tabular}{@{}c@{}}{\bf Src. Confid.}\end{tabular}} & 91.0 & 79.9 & 85.1 & 73.3 & 68.1 & 70.6\\\cline{2-8}
		&\multicolumn{1}{|c|}{\begin{tabular}{@{}c@{}}{\bf Pred. Score}\end{tabular}} & 92.8 & 82.8 & {\bf 87.5} & 76.6 & 70.1 & {\bf 73.2}\\
	\end{tabular}
	%\end{minipage}\hfill
	%\begin{minipage}[c]{0.36\textwidth}
	\caption{Test set Precision, Recall and F-measure scores for axiom parsing. These scores are computed over literals derived in axiom parses or full axiom parses. We show results for the old \textit{GEOS} system, for the improved \textit{GEOS++} system with expanded entity types, functions and predicates, and for the multi-source parsers presented in this paper.\label{parse:res}}
	%\end{minipage}
\end{table}
To evaluate axiom parsing, we compute precision, recall and F-score in (a) deriving literals in axiom parses, as well as for (b) the final axiom parses on our test set. Table \ref{parse:res} shows the results of axiom parsing for \textit{GEOS} (trained on the training set) as well as various versions of our best performing system (\textit{GEOS++} with our axiomatic solver) with various heuristics for multi-source parsing. The results show that our system (single source) performs better than \textit{GEOS} as it is trained with the expanded set of entity types, functions and predicates. The results also show that the choice of heuristic is important for the multi-source parser -- though all the heuristics lead to improvements over the single source parser. The average score heuristic that chooses the parse with the highest average score across sources performs better than majority voting which chooses the best parse based on a voting heuristic. Learning the confidence of every source and using a weighted average is an even better heuristic. Finally, predicate scoring which chooses the parse by scoring predicates on the premise and conclusion sides performs the best leading to 87.5 F1 score (when computed over parse literals) and 73.2 F1 score (when computed on the full parse). The high F1 score for axiom parsing on the test set shows that our approach works well and we can accurately harvest axiomatic knowledge from textbooks.

\begin{table}
	\center
	\begin{tabular}{c||c|c|c}
		& {\bf Practice} & {\bf Official} & {\bf Textbook} \tabularnewline
		\hline\hline
		\textit{GEOS} & 61 & 49 & 32 \\\hline
		{\bf Our System} & {\bf 64} & {\bf 55} & {\bf 51}\\\hline\hline
		{\bf Oracle} & 80 & 78 & 72 \\
	\end{tabular}
	\caption{Scores for solving geometry questions on the SAT practice and official datasets and a dataset of questions from the 20 textbooks. We use SAT's grading scheme that rewards a correct answer with a score of 1.0 and penalizes a wrong answer with a negative score of 0.25. \textit{Oracle} uses gold axioms but automatic text and diagram interpretation in our logical solver. All differences between \textit{GEOS} and our system are significant (p<0.05 using the two-tailed paired t-test).\label{solver:res}}
\end{table}
Finally, we use the extracted horn clause rules in our axiomatic solver for solving geometry problems.
For this, we over-generate a set of horn clause rules by generating 3 horn clause parses for each axiom and use them as the underlying theory in prolog programs such as the one shown in Figure \ref{fig:problog}. We use weighted logical expressions for the question description and the diagram derived from \textit{GEOS++} as declarations, and the (normalized) score of the parsing model multiplied by the score of the joint axiom identification and alignment model as weights for the rules.
Table \ref{solver:res} shows the results for our best end-to-end system and compares it to \textit{GEOS} on the practice and official SAT dataset from \citet{seo:2015} as well as questions from the 20 textbooks. On all the three datasets, our system outperforms \textit{GEOS}. Especially on the dataset from the 20 textbooks (which is indeed a harder dataset and includes more problems which require complex reasoning based on geometry), \textit{GEOS} doesn't perform very well whereas our system still achieves a good score. \textit{Oracle} shows the performance of our system when gold axioms (written down by an expert) are used along with automatic text and diagram interpretations in \textit{GEOS++}. This shows that there is scope for further improvement in our approach.
% and gives us hope that our approach can be used to help students learn geometry on forums such as MOOCs.

\section{Explainability}
Students around the world solve geometry problems through rigorous deduction whereas the numerical solver in \textit{GEOS} does not provide such explainability. One of the key benefits of our axiomatic solver is that it provides an easy-to-understand student-friendly deductive solution to geometry problems.
\begin{figure}
	\center
	\includegraphics[scale=0.4]{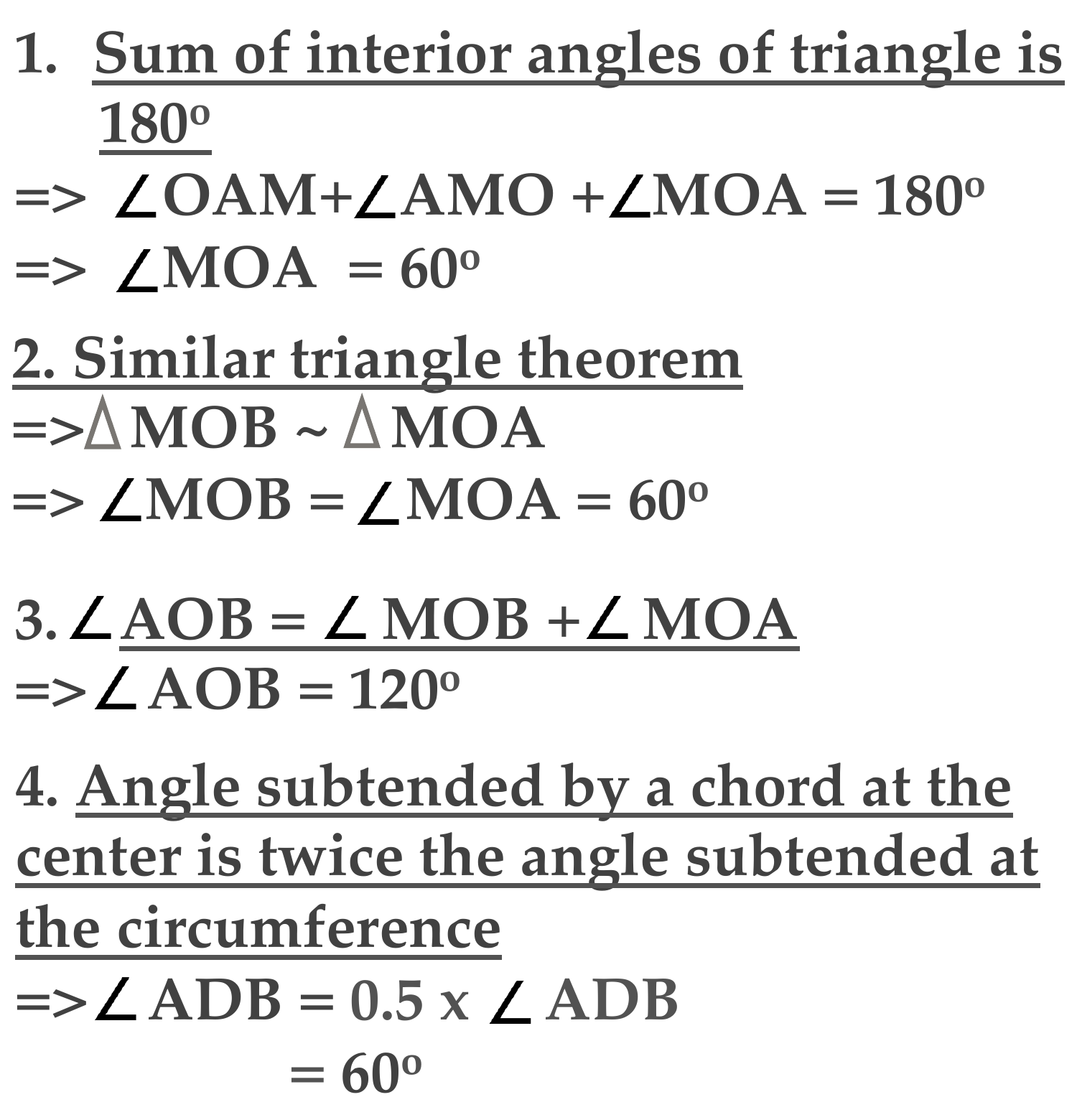}
	\caption{An example demonstration on how to solve the problem in Figure 1: (1) Use the theorem that the sum of interior angles of a triangle is 180\degree and additionally the fact that $\angle$AMO is 90\degree to conclude that $\angle$MOA is 60\degree. (2) Conclude that $\bigtriangleup$MOA $\sim$ $\bigtriangleup$MOB (using a similar triangle theorem) and then, conclude that $\angle$MOB = $\angle$MOA = 60\degree (using the theorem that corresponding angles of similar triangles are equal). (3) Use angle sum rule to conclude that $\angle$AOB = $\angle$MOB + $\angle$MOA = 120\degree. (4) Use the theorem that the angle subtended by an arc of a circle at the centre is double the angle subtended by it at any point on the circle to conclude that $\angle$ADB = 0.5$\times\angle$AOB = 60\degree.}\label{fig:answer}
\end{figure}

\begin{table}
	\center
	\begin{tabular}{c||c|c|c|c}
		& \multicolumn{2}{c}{{\bf Explainability}} & \multicolumn{2}{|c}{{\bf Usefulness}}\\\hline
		& {\bf \textit{GEOS}} & {\bf \textit{O.S.}} & {\bf \textit{GEOS}} & {\bf \textit{O.S.}} \tabularnewline
		\hline\hline
		{\bf Grade 6} & 2.7 & {\bf 2.9} & 2.9 & {\bf 3.2} \\\hline
		{\bf Grade 7} & 3.0 & {\bf 3.7} & 3.3 & {\bf 3.6}\\\hline
		{\bf Grade 8} & 2.7 & {\bf 3.5} & 3.1 & {\bf 3.5}\\\hline
		{\bf Grade 9} & 2.4 & {\bf 3.3} & 3.0 & {\bf 3.7}\\\hline
		{\bf Grade 10} & 2.8 & {\bf 3.1} & 3.2 & {\bf 3.8}\\\hline\hline
		{\bf Overall} & 2.7 & {\bf 3.3} & 3.1 & {\bf 3.6}\\
	\end{tabular}
	\caption{User study ratings for \textit{GEOS} and our system (O.S.) by students in grade 6-10. Ten students in each grade were asked to rate the two systems on a scale of 1-5 on two facets: `explainability' and `usefulness'. Each cell shows the mean rating computed over ten students in that grade for that facet.\label{rating:res}}
\end{table}
To test the explainability of our axiomatic solver, we asked 50 grade 6-10 students (10 students in each grade) to use \textit{GEOS} and our system (\textit{GEOS++} with our axiomatic solver) as a web-based assistive tool while learning geometry.
The tool uses the probabilistic prolog solver \cite{fierens:2015} to derive the most probable explaination (MPE) for a solution. Then, it lists, one by one, the various axioms used and the conclusion drawn from the axiom application as shown in Figure \ref{fig:answer}.
The students were each asked to rate how `explainable' and `useful' the two systems were on a scale of 1-5. Table \ref{rating:res} shows the mean rating by students in each grade on the two facets. We can observe that students of each grade found our system to be more interpretable as well as more useful to them than \textit{GEOS}. This study lends support to our claims about the need of an interpretable deductive solver for geometry problems.

\section{Feature Ablation}\label{sec:feats}
In this section, we will measure the value of the various features in our axiom harvesting and parsing pipeline. Note that we have described three set of features ${\bf f}$, ${\bf g}$ and ${\bf h}$ corresponding to the various steps in our pipeline: axiom identification, axiom alignment and axiom parsing in Tables \ref{tab:feats1}, \ref{tab:feats2} and \ref{tab:feats3}. We will ablate each of the three features one by one via \textit{backward selection}, i.e. we will remove features and observe how that affects performance.

\subsection{Ablating Axiom Identification Features}
	\begin{table}%[h]
	\newcommand{\ra}[1]{\renewcommand{\arraystretch}{#1}}
	\ra{.6}
	\centering
	%\begin{tabular}{@{} L{20mm} R{20mm} R{20mm} R{20mm} R{20mm} R{20mm} @{}} 
	%\begin{tabular}{@{} l r  r r r r  r r r  r@{}} 
	% 		\resizebox{\textwidth}{!}{%
	\begin{tabular}{@{} l c c c c c c c@{}} 
		\toprule
		& & \multicolumn{2}{c}{\small \textsf{Axiom Identification F1}} & \phantom{a}  & \multicolumn{3}{c}{\small \textsf{SAT Scores}} \\
		\cmidrule{3-4} \cmidrule{6-8}
		
		& & {\small \textsf{Strict Comp.}} & {\small \textsf{Relaxed Comp.}} & \phantom{a} & {\small \textsf{Practice}} & {\small \textsf{Official}} & {\small \textsf{Textbook}} \\
		\midrule
		\parbox[t]{1mm}{\multirow{3}{*}{\rotatebox[origin=c]{90}{{\scriptsize Content}}}} & {\small  \textsf{Sentence Overlap}} & 56.2 & 73.8 &~ & 56 & 43 & 42\\
		&{\small  \textsf{Geometry entities}} & 64.0 & 80.4 &~ & 61 & 49 & 46\\
		&{\small  \textsf{Keywords}} & 67.5 & 81.0 &~ & 62 & 54 & 48\\
		\midrule
		\parbox[t]{1mm}{\multirow{8}{*}{\rotatebox[origin=c]{90}{{\scriptsize Discourse (Typography)}}}} 		&{\small  \textsf{RST edge}} & 66.6 & 78.9 &~ & 58 & 46 & 44\\
		& {\small  \textsf{Axm, Thm, Corr.}} & 62.6 & 77.8 &~ & 57 & 47 & 43\\
		&{\small  \textsf{Equation}} & 66.2 & 78.6 &~ & 57 & 46 & 42\\
		&{\small  \textsf{Associated Diagram}} & 68.5 & 84.4 &~ & 61 & 52 & 49\\
		&{\small  \textsf{Bold / Underline}} & 68.2 & 82.0 &~ & 62 & 52 & 48\\
		&{\small  \textsf{Bounding box}} & 59.7 & 75.5 &~ & 55 & 47 & 40\\
		&{\small  \textsf{XML structure}} & 67.4 & 80.6 &~ & 60 & 51 & 46\\
		\midrule
		&{\small  \textsf{Unablated}} & 70.4 & 87.6 &~ &64 & 55 & 51\\
		\bottomrule
	\end{tabular}
	\caption{Ablation study results for the axiom identification component. We remove features of the axiom identification component one by one as listed in Table \ref{tab:feats1} and observe the fall in performance in terms of the axiom identification performance as well as the overall performance to gauge the value of the various features.}\label{ablate:1}
	% 		}
\end{table}
Table \ref{ablate:1} shows the fall in performance in terms of the axiom identification performance as well as the overall performance as we ablate various axiom identification features listed in Table \ref{tab:feats1}. We can observe that removal of any of the features results in a loss of performance. Thus, all the content as well as typographical features are important for performance. We observe that the content features such as sentence overlap, geometry entity sharing and keyword usage are clearly important. At the same time, the various discourse features such as the RST relation, axiom, theorem, corollary annotation, use of equations and diagrams, bold/underline, bounding box and XML structure are all important. Most of these features depend on typographical information which vital in performance of the axiom identification component as well as the overall model. In particular, we can observe that the axiom, theorem, corollary annotation and the bounding box features contribute most to the performance of the model as they are direct indicators of the presence of an axiom mention.

\subsection{Ablating Axiom Alignment Features}
	\begin{table}%[h]
	\newcommand{\ra}[1]{\renewcommand{\arraystretch}{#1}}
	\ra{.6}
	\centering
	%\begin{tabular}{@{} L{20mm} R{20mm} R{20mm} R{20mm} R{20mm} R{20mm} @{}} 
	%\begin{tabular}{@{} l r  r r r r  r r r  r@{}} 
	% 		\resizebox{\textwidth}{!}{%
	\begin{tabular}{@{} l c c c c c c c@{}} 
		\toprule
		& & \multicolumn{2}{c}{\small \textsf{}} & \phantom{a}  & \multicolumn{3}{c}{\small \textsf{SAT Scores}} \\
		\cmidrule{6-8}
		
		& & {\small \textsf{F1}} & {\small \textsf{NMI}} & \phantom{a} & {\small \textsf{Practice}} & {\small \textsf{Official}} & {\small \textsf{Textbook}} \\
		\midrule
		\parbox[t]{1mm}{\multirow{7}{*}{\rotatebox[origin=c]{90}{{\scriptsize Content}}}} & {\small  \textsf{Overlap}} & 70.7 & 0.54 &~ & 57 & 45 & 45\\
		&{\small  \textsf{LCS}} & 78.7 & 0.64 &~ & 61 & 53 & 49\\
		&{\small  \textsf{Number of Sentences}} & 78.5 & 0.65 &~ & 62 & 54 & 48\\
		& {\small  \textsf{Alignment Scores}} & 72.6 & 0.57 &~ & 59 & 49 & 48\\
		&{\small  \textsf{MT Metrics}} & 74.8 & 0.60 &~ & 62 & 52 & 49\\
		&{\small  \textsf{Summarization Metrics}} & 75.9 & 0.63 &~ & 62 & 54 & 50\\
		\midrule
		\parbox[t]{1mm}{\multirow{3}{*}{\rotatebox[origin=c]{90}{{\tiny Typography}}}} &{\small  \textsf{XML Structure}} & 71.5 & 0.57 &~ & 58 & 47 & 46\\
		&{\small  \textsf{Equation Template}} & 76.6 & 0.61 &~ & 57 & 47 & 43\\
		&{\small  \textsf{Image Caption}} & 77.9 & 0.65 &~ & 62 & 53 & 47\\
		\midrule
		&{\small  \textsf{Unablated}} & 80.3 & 0.69 &~ &64 & 55 & 51\\
		\bottomrule
	\end{tabular}
	\caption{Ablation study results for the axiom alignment component. We remove features of the axiom alignment component one by one as listed in Table \ref{tab:feats2} and observe the fall in performance in terms of the axiom alignment performance as well as the overall performance to gauge the value of the various features.}\label{ablate:2}
	% 		}
\end{table}
Table \ref{ablate:2} shows the fall in performance in terms of the axiom alignment performance as well as the overall performance as we ablate various axiom alignment features listed in Table \ref{tab:feats2}. We again observe that removal of any of the features results in a loss of performance. Thus, the various content as well as typographical features are important for performance. We observe that the content features such as unigram, bigram and entity overlap, length of the longest common subsequence, number of sentences and various aligner, MT and summarization scores are clearly important. At the same time, the various discourse features such as the XML structure, equation template and image caption match are all important. Note that these features depend on typographical information which is again vital in performance. In particular, we can observe that the overlap and the XML structure features contribute most to the performance of the model.

\subsection{Ablating Axiom Parsing Features}
	\begin{table}%[h]
	\newcommand{\ra}[1]{\renewcommand{\arraystretch}{#1}}
	\ra{.6}
	\centering
	%\begin{tabular}{@{} L{20mm} R{20mm} R{20mm} R{20mm} R{20mm} R{20mm} @{}} 
	%\begin{tabular}{@{} l r  r r r r  r r r  r@{}} 
	% 		\resizebox{\textwidth}{!}{%
	\begin{tabular}{@{} l c c c c c c c@{}} 
		\toprule
		& & \multicolumn{2}{c}{\small \textsf{F1}} & \phantom{a}  & \multicolumn{3}{c}{\small \textsf{SAT Scores}} \\
		\cmidrule{3-4} \cmidrule{6-8}
		
		& & {\small \textsf{Literals}} & {\small \textsf{Full Parse}} & \phantom{a} & {\small \textsf{Practice}} & {\small \textsf{Official}} & {\small \textsf{Textbook}} \\
		\midrule
		&{\small  \textsf{Span Similarity}} & 71.8 & 64.6 &~ & 51 & 40 & 42\\
		&{\small  \textsf{No. of Relations}} & 82.3 & 70.5 &~ & 60 & 51 & 49\\
		&{\small  \textsf{Span Lengths}} & 86.0 & 72.0 &~ & 63 & 54 & 50\\
		&{\small  \textsf{Relative Position}} & 83.9 & 69.2 &~ & 60 & 52 & 47\\
		&{\small  \textsf{Discourse Markers}} & 77.4 & 68.4 &~ & 55 & 48 & 47\\
		&{\small  \textsf{Punctuations}} & 73.5 & 65.0 &~ & 52 & 45 & 45\\
		& {\small  \textsf{Text Organization}} & 74.4 & 66.2 &~ & 52 & 47 & 46\\
		&{\small  \textsf{RST Parse}} & 84.6 & 70.8 &~ & 62 & 52 & 49\\
		&{\small  \textsf{Soricut \& Marcu}} & 83.2 & 69.8 &~ & 61 & 52 & 50\\
		&{\small  \textsf{Head Node, etc.}} & 85.3 & 71.6 &~ & 62 & 54 & 49\\
		&{\small  \textsf{Syntax}} & 75.5 & 66.6 &~ & 54 & 47 & 46\\
		&{\small  \textsf{Dominance}} & 73.9 & 66.1 &~ & 53 & 47 & 44\\
		&{\small  \textsf{XML Structure}} & 77.6 & 68.0 &~ & 59 & 51 & 46\\
		\midrule
		&{\small  \textsf{Unablated}} & 87.5 & 73.2 &~ &64 & 55 & 51\\
		\bottomrule
	\end{tabular}
	\caption{Ablation study results for the axiom parsing component. We remove features of the axiom parsing component one by one as listed in Table \ref{tab:feats3} and observe the fall in performance in terms of the axiom parsing performance as well as the overall performance to gauge the value of the various features.}\label{ablate:3}
	% 		}
\end{table}
Table \ref{ablate:3} shows the fall in performance in terms of the axiom parsing performance as well as the overall performance as we ablate various axiom parsing features listed in Table \ref{tab:feats3}. We again observe that removal of any of the features results in a loss of performance. The axiom parsing component uses few content based features such as span similarity and  no. of relations, span lengths and relative position, and various discourse features such as discourse markers, punctuations, text organization, RST parse, an existing discourse segmentor from Soricut and Marcu \cite{soricut:2003}, node attachment, syntax, dominance and XML structure, and all are clearly important. In particular, we can observe that span similarity and punctuation features contribute most to the performance of the model.

\section{Axioms Harvested}
\begin{figure}
	\center
	\includegraphics[scale=0.4]{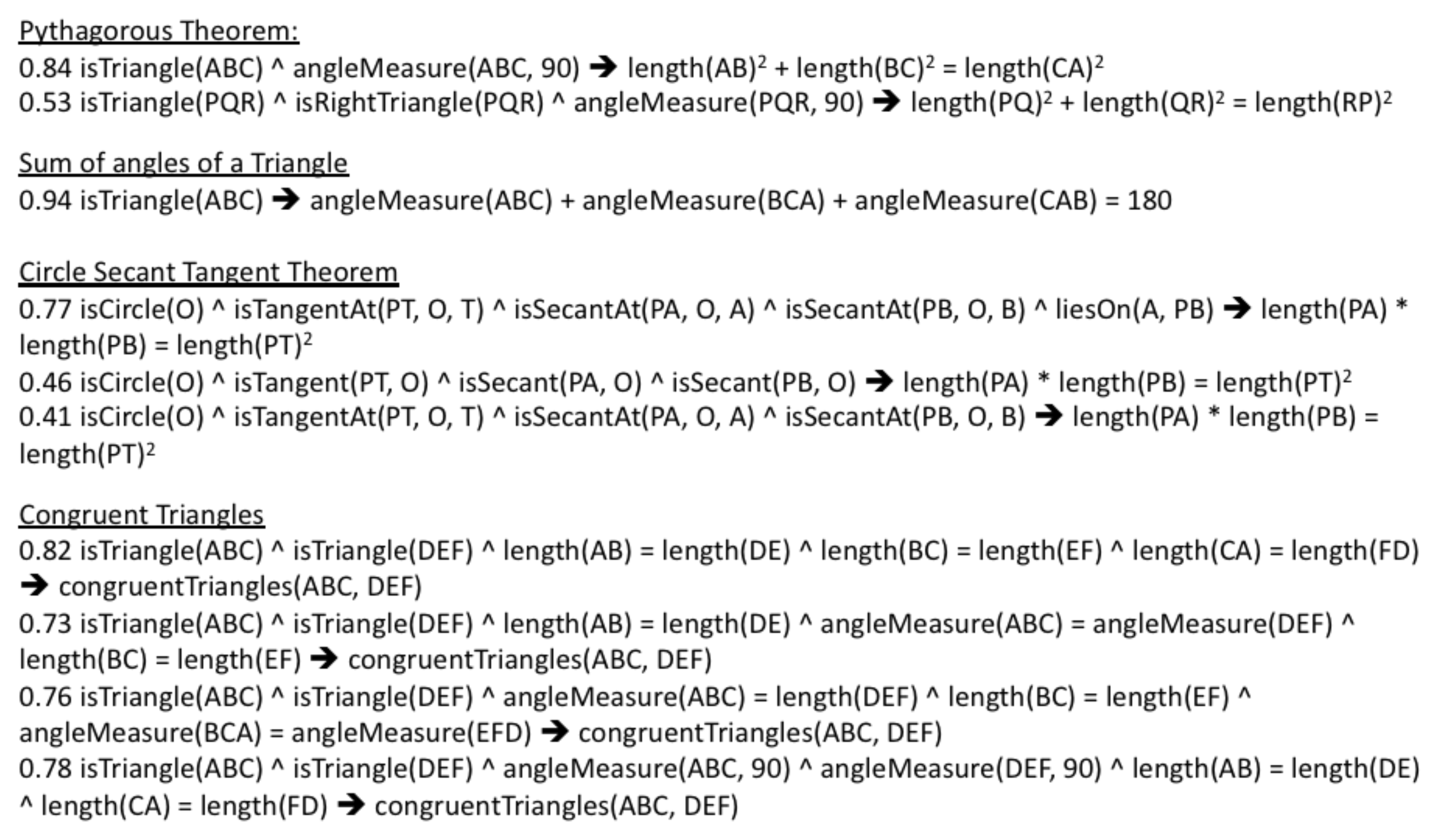}
	\caption{Horn clause rules for some popular named theorems in geometry harvested by our approach. We also show the confidence our method has on the rule being correct (which is used in reasoning via the problog solver).}\label{fig:theorems}
\end{figure}
We qualitatively analyze the structured axioms harvested by our method. We show few most probable horn clause rules for some popular named theorems in geometry in Figure \ref{fig:theorems} along with the confidence of our method on the rules being correct. Note that some horn clause parsed rules can be incorrect. For example, the second most probable horn clause rule for the Pythagoras theorem is partially incorrect (doesn't state which angle is 90\degree). Similarly, the second and third most probable horn clause for the circle secant tangent theorem are also incorrect. Our problog solver can use these redundant but weighted horn clause rules for solving geometry problems.

\section{Example Solutions and Error Analysis}
\begin{table}\center
	\begin{tabular}{|c|c|}\hline
		\includegraphics[scale=0.25]{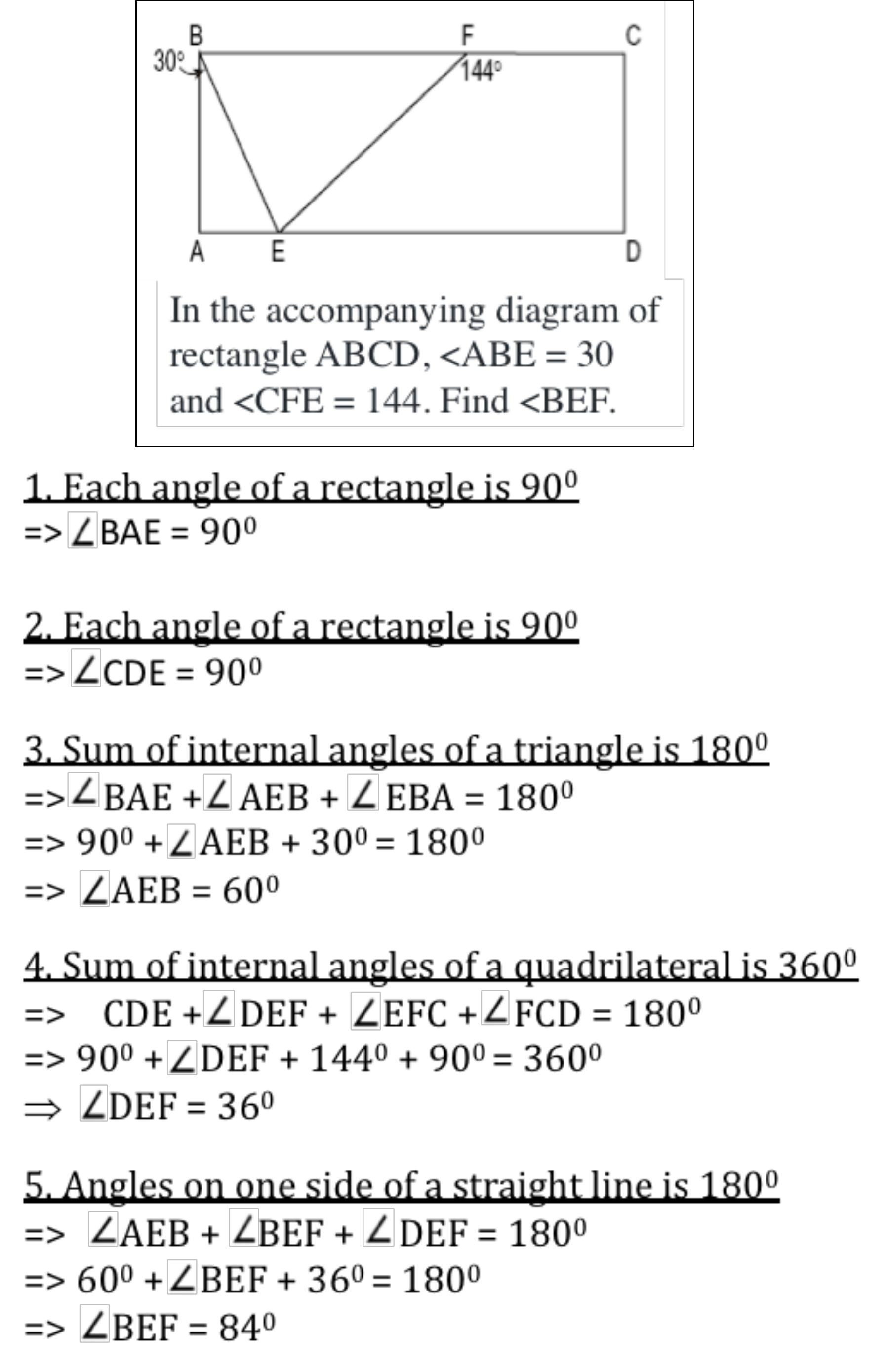} & \includegraphics[scale=0.3]{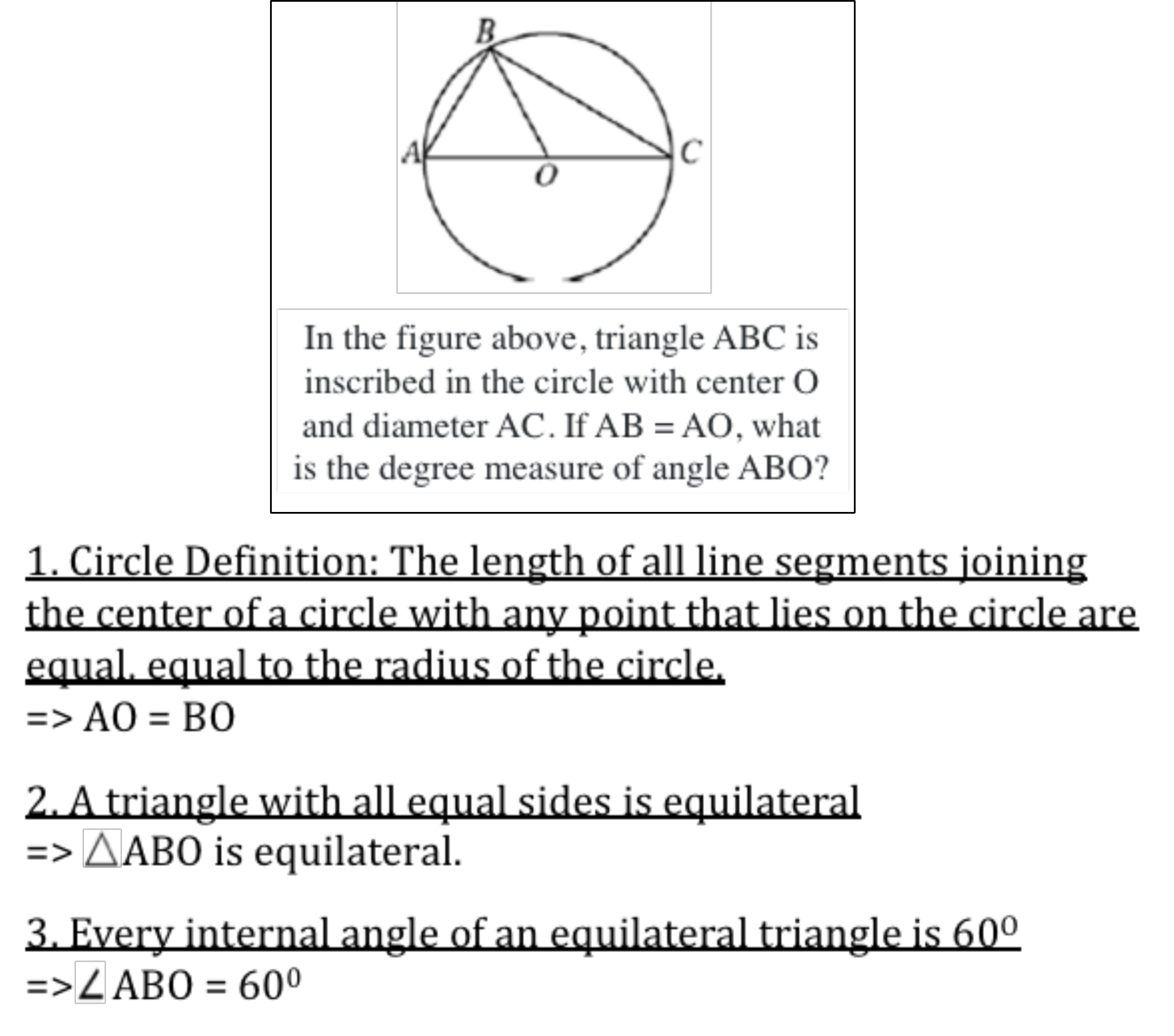}\\\hline
		\includegraphics[scale=0.3]{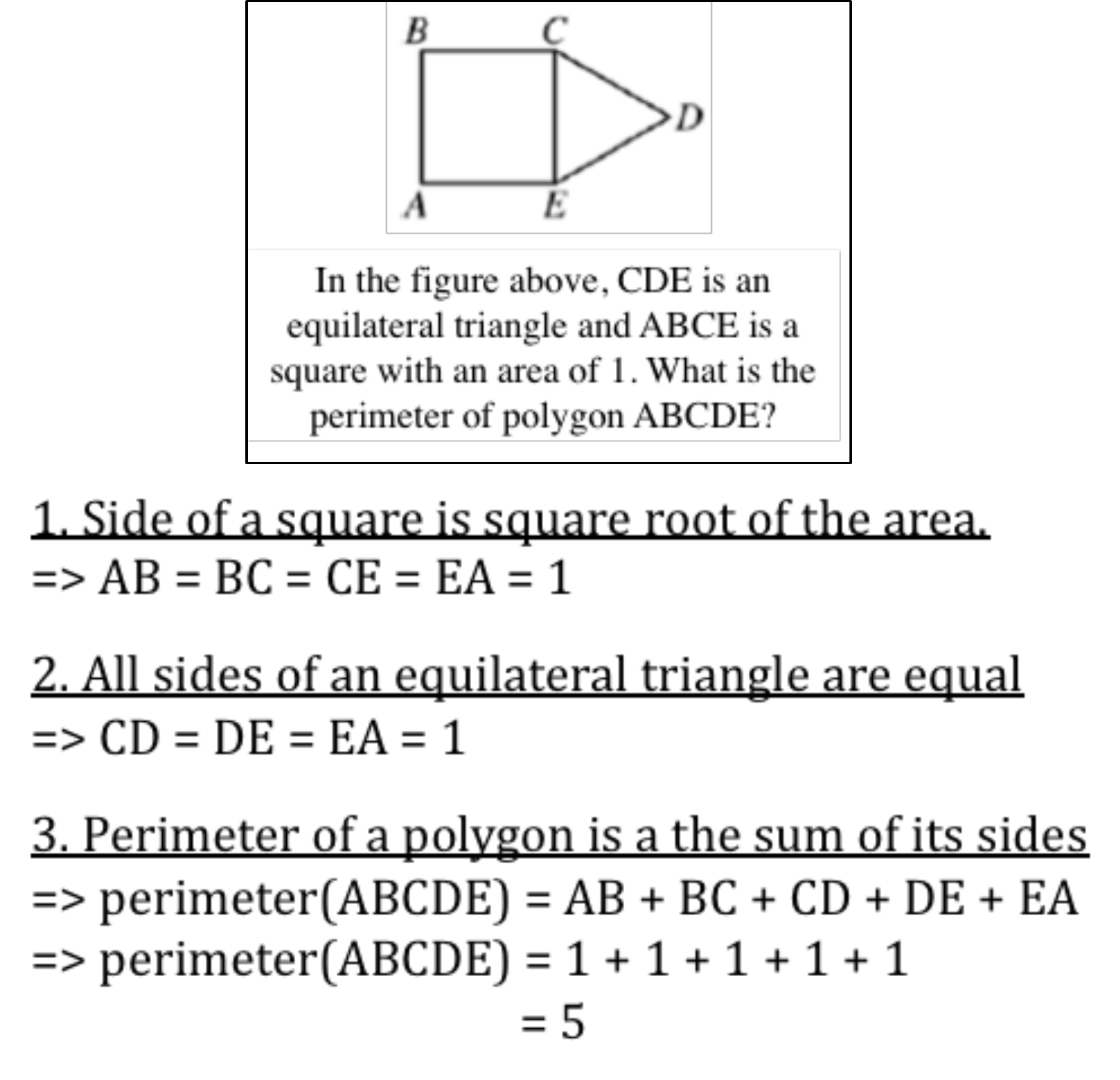} & \includegraphics[scale=0.3]{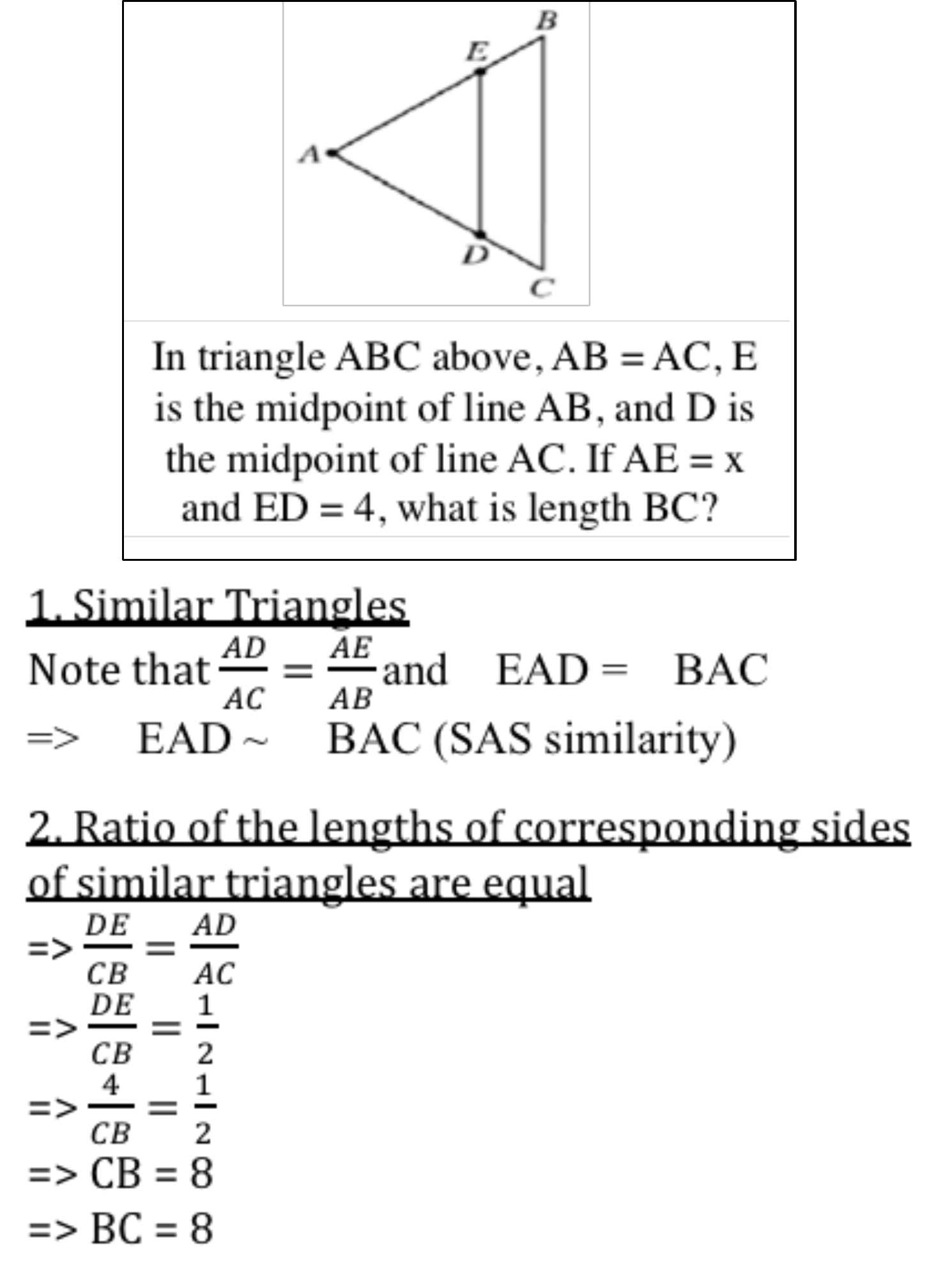}\\\hline
	\end{tabular}
	\caption{Some correctly answered questions along with explanations generated by our deductive solver for these problems.}\label{fig:answer1}
\end{table}
Next, we qualitatively describe some example solutions of geometry problems as well as perform a qualitative error analysis. We first show some sample questions which our solver can answer correctly in Table \ref{fig:answer1}. We also show the explanations generated by our deductive solver for these problems (constructed in the same way as described earlier). Note that these problems are diverse in terms of question types as well as the reasoning required to answer them and our solver can handle them.

We also show some failure cases of our approach in Table \ref{fig:answer2}. There are a number of reasons that could lead to a failure of our approach to correctly answer a question. These include an error in parsing the diagram, the text, or an incorrect or incomplete knowledge in the form of geometry rules. As can be observed in the failure examples, and also evaluated by us in a small error analysis of 100 textbook questions, our approach answered 52 questions correctly. Amongst the 48 incorrectly answered questions, our diagram parse was incorrect for 12 questions, and the text parse was incorrect for 15 questions. Our formal language was insufficiently defined to handle 6 questions, i.e. the semantics of the question could not be adequately captured by the formal language. 21 questions were incorrectly answered due to missing knowledge of geometry in the form of rules. Note that several questions were incorrectly answered due to a failure of multiple system components (for example, failure of both the text and the diagram parser).
\begin{table}\center
	\begin{tabular}{|c|c|}\hline
		\includegraphics[scale=0.275]{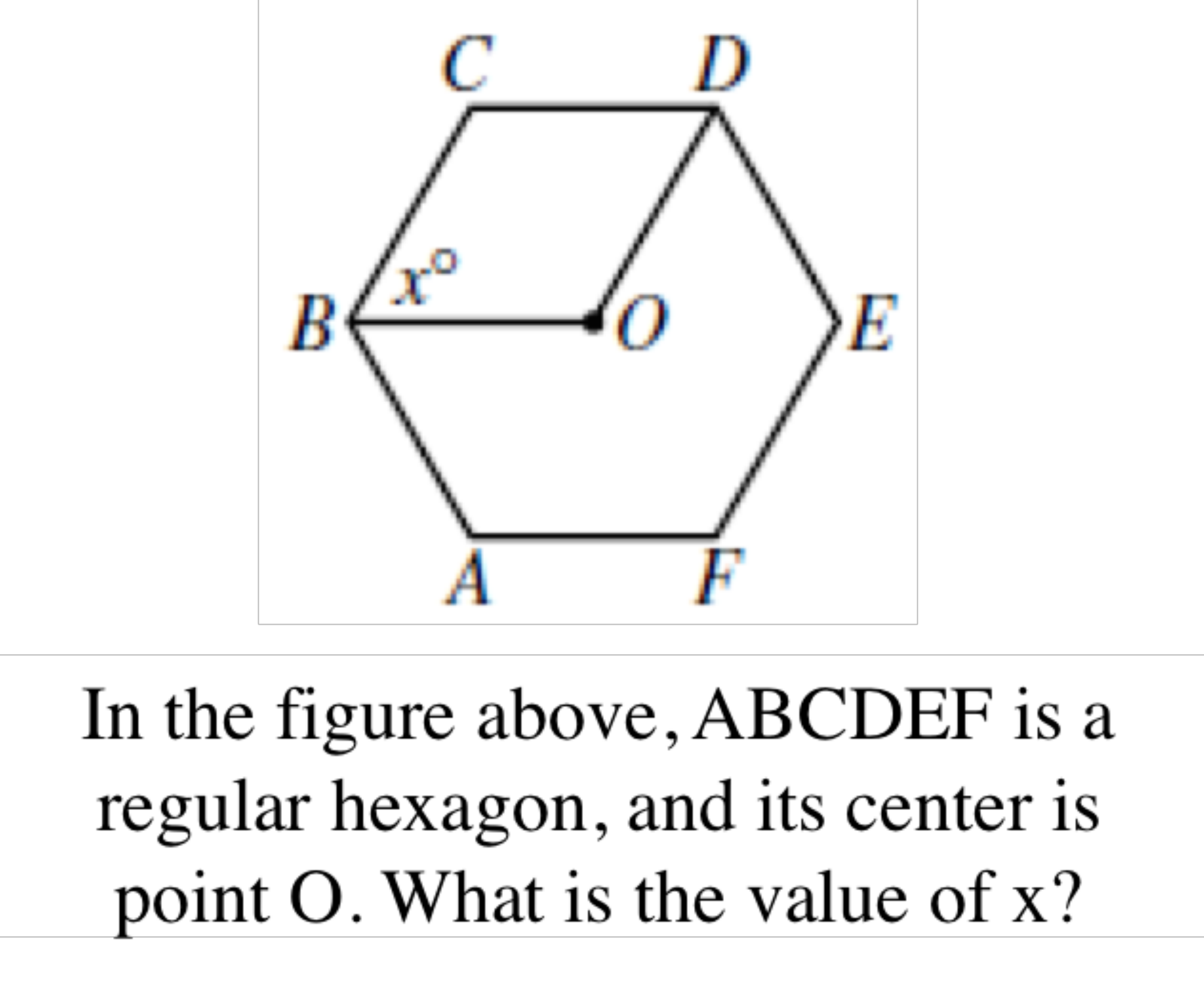} & \includegraphics[scale=0.275]{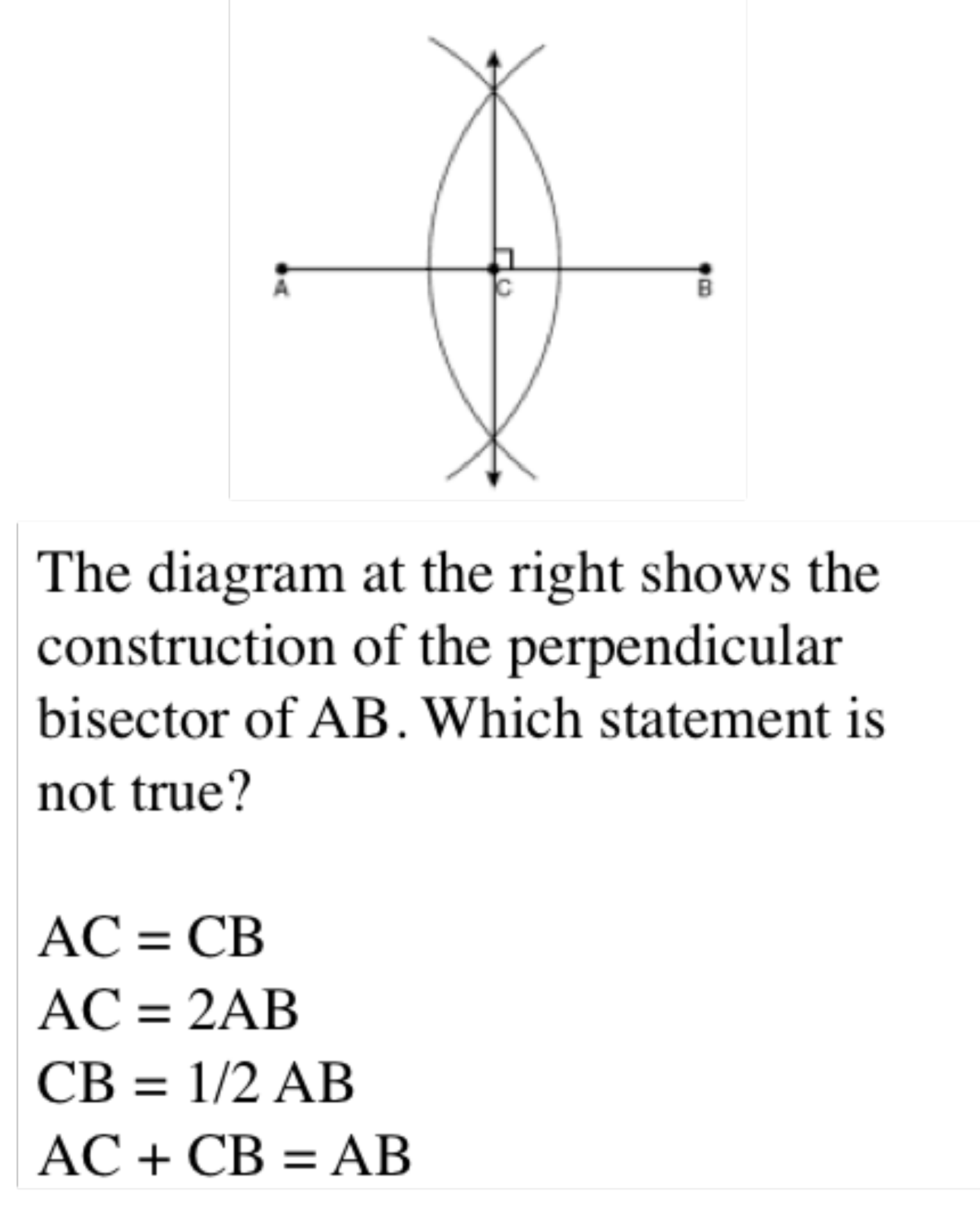}\\\hline
		\includegraphics[scale=0.275]{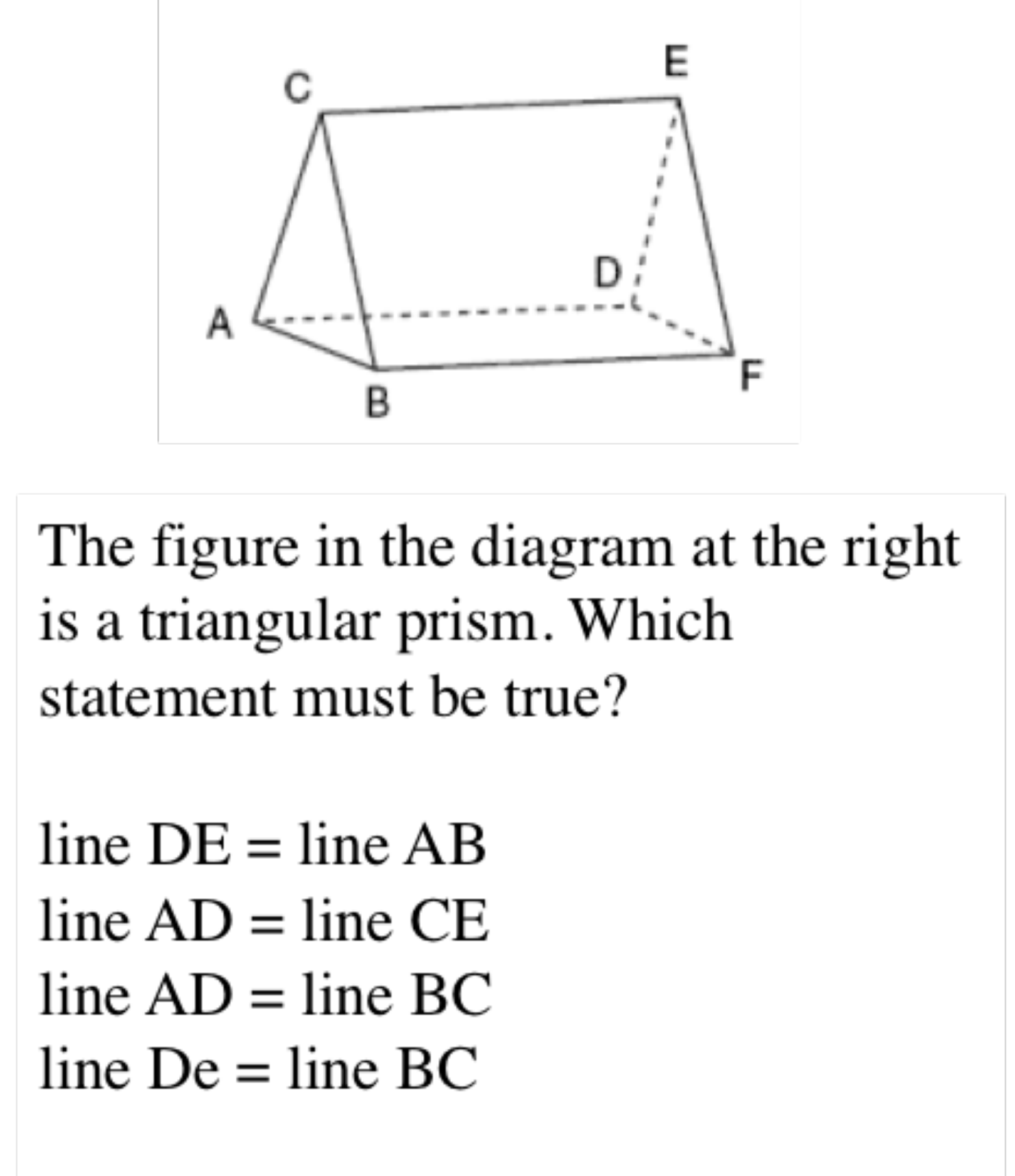} & \includegraphics[scale=0.275]{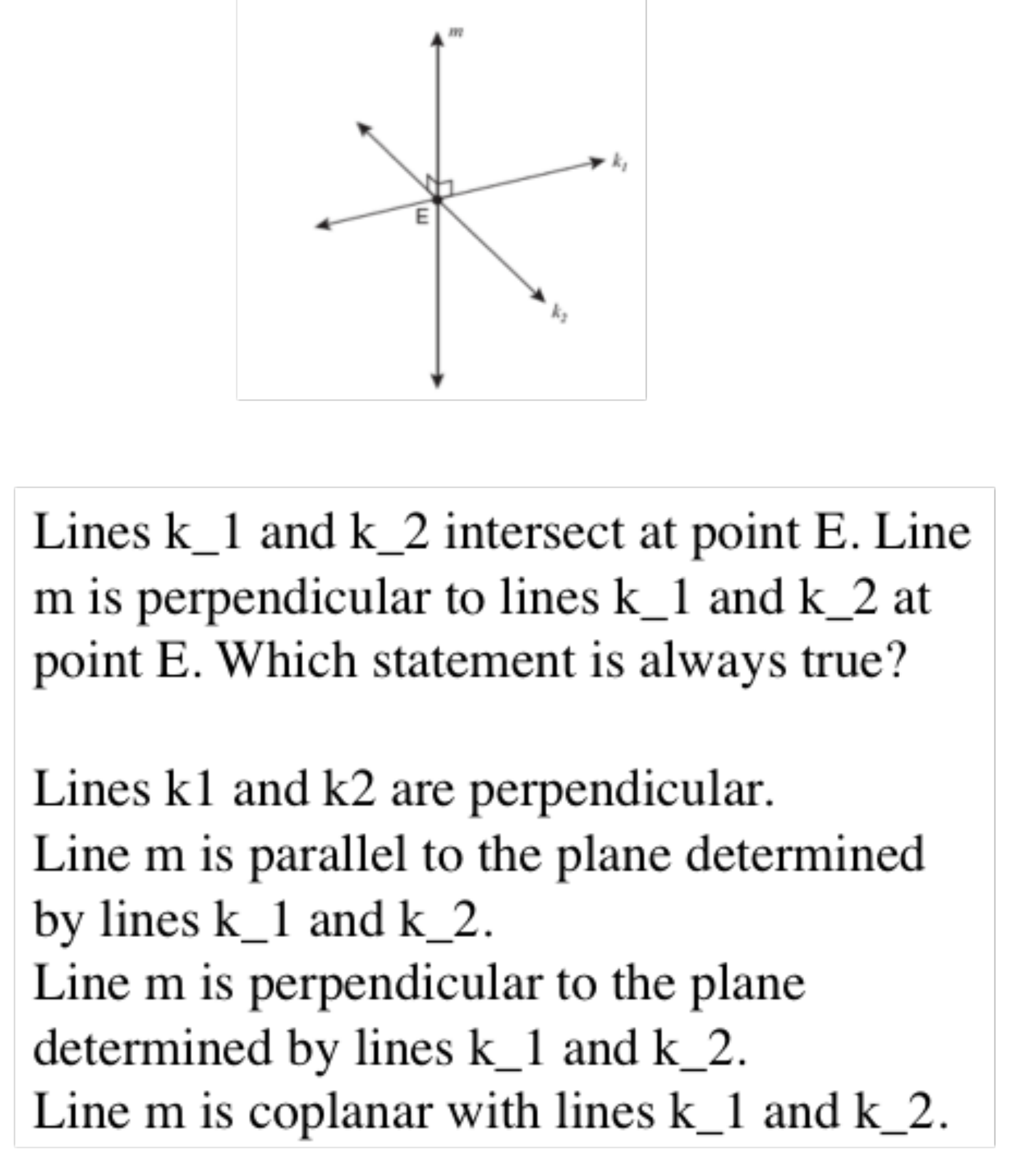}\\\hline
		\includegraphics[scale=0.275]{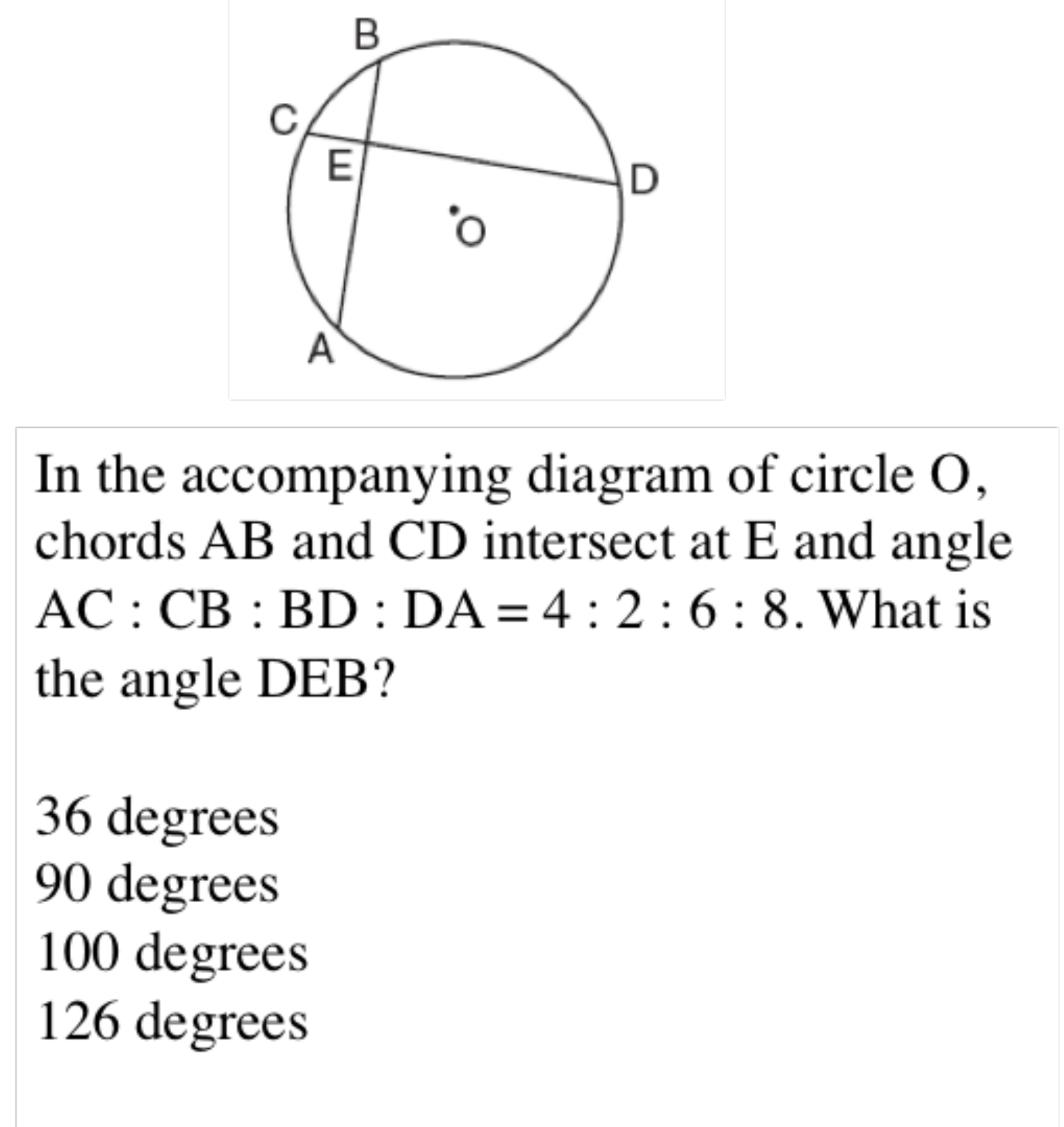} & \includegraphics[scale=0.275]{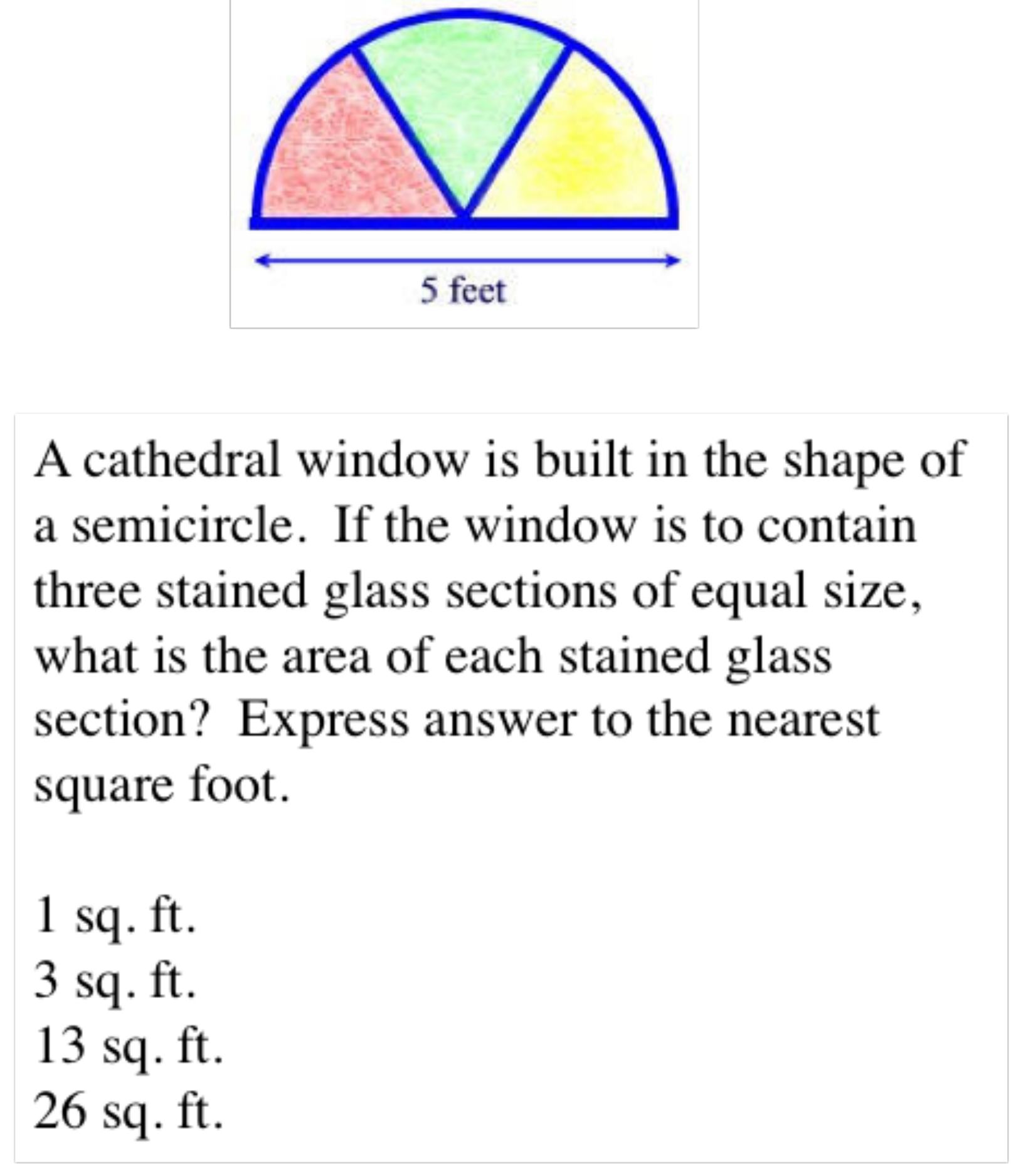}\\\hline
	\end{tabular}
	\caption{Some example failure cases of our approach for solving SAT style geometry problems. In (i) the axiom set contains an axiom that internal angle of a regular hexagon is 120\degree and that each side of a regular polygon is equal. But there's no way to deduce that the angle CBO is half of the internal angle ABC (by symmetry).
The other hand, the coordinate geometry solver can exploit these three facts as maximizing the satisfiability of the various constraints can answer the question.
(ii) The solver does not contain any knowledge about construction. The question cannot be correctly interpreted and the coordinate geometry solver also gets it wrong.
(iii) The solver does not contain any knowledge about construction or prisms. The question cannot be correctly interpreted and the coordinate geometry solver also gets it wrong.
(iv) The question as well as the answer candidates cannot be correctly interpreted (as the concept of perpendicular to plane is not in the vocabulary). Both solvers get it wrong.
(v) The parser cannot interpret that angle AC is indeed angle AEC. This needs to be understood by context as it defies the standard type definition of an angle. Both solvers get it wrong. (vi) Both diagram and text parsers fail here. Both solvers answer incorrectly.
	}\label{fig:answer2}
\end{table}

\section{Conclusion}
We presented an approach to harvest structured axiomatic knowledge from math textbooks. Our approach uses rich features based on context and typography, the redundancy of axiomatic knowledge and shared ordering constraints across multiple textbooks to accurately extract and parse axiomatic knowledge to horn clause rules. We used the parsed axiomatic knowledge to improve the best previously published automatic approach to solve geometry problems. A user-study conducted on a number of school students studying geometry found our approach to be more interpretable and useful than its predecessor. While this paper focused on harvesting geometry axioms from textbooks as a case study, we believe that it can be extended to obtain valuable structured knowledge from textbooks in areas such as science, engineering and finance.

\starttwocolumn
\bibliography{compling_style}
\bibliographystyle{compling}

\end{document}